\documentclass[acmlarge]{acmart}
\usepackage{tabularx}
\usepackage{array}
\usepackage{multicol}
\usepackage{multirow}
\usepackage{makecell}

\renewcommand\footnotetextcopyrightpermission[1]{} 
\usepackage{pifont}
\newcommand{\cmark}{\ding{51}}%
\newcommand{\xmark}{\ding{55}}%

\newcolumntype{T}[1]{>{\centering\arraybackslash}p{#1}}
\usepackage[parfill]{parskip}    

\makeatletter
\def\subsubsection{\@startsection{subsubsection}{3}%
  \z@{.5\linespacing\@plus.7\linespacing}{.1\linespacing}%
  {\normalfont\itshape}}
\makeatother

\AtBeginDocument{%
 \providecommand\BibTeX{{%
 \normalfont B\kern-0.5em{\scshape i\kern-0.25em b}\kern-0.8em\TeX}}}

\setcopyright{acmcopyright}
\copyrightyear{2021}
\acmYear{2021}
\acmDOI{0}

\acmJournal{JACM}
\acmVolume{0}
\acmNumber{0}
\acmArticle{0}
\acmMonth{0}

\begin{document}

\title{Machine Learning-based Orchestration of Containers: A Taxonomy and Future Directions}

\author{Zhiheng Zhong}
\affiliation{%
 \institution{The University of Melbourne}
 \streetaddress{Grattan Street, Parkville}
 \city{Melbourne}
 \state{Victoria}
 \country{Australia}
 \postcode{VIC 3010}
}

\author{Minxian Xu}
\affiliation{%
 \institution{Shenzhen Institute of Advanced Technology, Chinese Academy of Sciences}
 \streetaddress{1068 Xueyuan Avenue, Shenzhen University Town,}
 \city{Shenzhen}
 \state{Guangdong}
 \country{China}
 \postcode{518055}
}

\author{Maria Alejandra Rodriguez}
\affiliation{%
 \institution{The University of Melbourne}
 \streetaddress{Grattan Street, Parkville}
 \city{Melbourne}
 \state{Victoria}
 \country{Australia}
 \postcode{VIC 3010}
}

\author{Chengzhong Xu}
\affiliation{%
 \institution{The University of Macau}
 \streetaddress{Avenida da Universidade}
 \city{Taipa}
 \state{Macau}
 \country{China}
 \postcode{999078}
}

\author{Rajkumar Buyya}
\affiliation{%
 \institution{The University of Melbourne}
 \streetaddress{Grattan Street, Parkville}
 \city{Melbourne}
 \state{Victoria}
 \country{Australia}
 \postcode{VIC 3010}
}

\authorsaddresses{%
Authors’ addresses: Zhiheng Zhong, Maria Alejandra Rodriguez, and Rajkumar Buyya, The Cloud Computing and Distributed Systems Laboratory, School of Computing and Information Systems, The University of Melbourne, Grattan Street, Parkville, Melbourne, Victoria, Australia, VIC 3010; emails: zhiheng@student.unimelb.edu.au, maria.read@unimelb.edu.au, rbuyya@unimelb.edu.au; Minxian Xu (Corresponding Author), Shenzhen Institute of Advanced Technology, Chinese Academy of Sciences, 1068 Xueyuan Avenue, Shenzhen University Town, Shenzhen, Guangdong, China, 518055; email: mx.xu@siat.ac.cn; Chengzhong Xu, State Key Lab of IOTSC, The University of Macau, Avenida da Universidade, Taipa, Macau, China, 999078; email: czxu@um.edu.mo}

\renewcommand{\shortauthors}{Zhiheng Zhong, Minxian Xu, Maria Alejandra Rodriguez, Chengzhong Xu, and Rajkumar Buyya}

\begin{abstract}
 Containerization is a lightweight application virtualization technology, providing high environmental consistency, operating system distribution portability, and resource isolation. Existing mainstream cloud service providers have prevalently adopted container technologies in their distributed system infrastructures for automated application management. To handle the automation of deployment, maintenance, autoscaling, and networking of containerized applications, container orchestration is proposed as an essential research problem. However, the highly dynamic and diverse feature of cloud workloads and environments considerably raises the complexity of orchestration mechanisms. Machine learning algorithms are accordingly employed by container orchestration systems for behavior modelling and prediction of multi-dimensional performance metrics. Such insights could further improve the quality of resource provisioning decisions in response to the changing workloads under complex environments. In this paper, we present a comprehensive literature review of existing machine learning-based container orchestration approaches. Detailed taxonomies are proposed to classify the current researches by their common features. Moreover, the evolution of machine learning-based container orchestration technologies from the year 2016 to 2021 has been designed based on objectives and metrics. A comparative analysis of the reviewed techniques is conducted according to the proposed taxonomies, with emphasis on their key characteristics. Finally, various open research challenges and potential future directions are highlighted.

 \end{abstract}


\maketitle

\section{Introduction}
\label{section:intro}

Our era has witnessed the trend of cloud computing becoming the mainstream industrial computing paradigm, providing stable service delivery with high cost-efficiency, scalability, availability, and accessibility  \cite{cloudcom}. Existing mainstream cloud service providers, including Amazon Web Services (AWS) \cite{aws}, Google \cite{borg}, and Alibaba \cite{alibabasemi}, prevalently adopt virtualization technologies including virtual machines (VM) and containers in their distributed system infrastructures for automated application deployment. In recent years, their infrastructures are evolving from VM centric to container centric \cite{soatcontainertech}. 

Containers employ a logical packing mechanism that binds both software and dependencies together for application abstraction \cite{costeffi}. Unlike VMs that support hardware-level resource virtualization where each VM has to maintain its own operating system (OS), containers virtualize resources at the OS level where all the containers share the same OS with less overhead. Therefore, containers are more lightweight in nature with high application portability, resource efficiency, and environmental consistency. They define a standardized unit for application deployment under an isolated runtime system. Thanks to these features, we have observed the prevalence of container technologies for automatic application deployment under diverse cloud environments.

Following this trend of containerization,  container orchestration techniques are proposed for the management of containerized applications. Container orchestration is the automated management process of a container lifecycle, including resource allocation, deployment, autoscaling, health monitoring, migration, load balance, security, and network configuration. For cloud service providers which have to handle hundreds or thousands of containers simultaneously, a refined and robust container orchestration system is the key factor in controlling overall resource utilization, energy efficiency, and application performance. Under the surge of cloud workloads in terms of resource demands, bandwidth consumption, and quality of service (QoS) requirements, the traditional cloud computing environment is extended to fog and edge infrastructures that are close to end users with extra computational power \cite{edgecomputing}. Consequently, this also requires current container orchestration systems to be further enhanced in response to the rising resource heterogeneity, application distribution, workload diversity, and security requirements across hybrid cloud infrastructures.

\subsection{Needs for Machine Learning-based Container Orchestration}

Considering the increasingly diverse and dynamic cloud workloads processed by existing cloud service providers, it is still unclear how to automate the orchestration process for complex heterogeneous workloads under large-scale cloud computing systems \cite{htas,microservicereview}. 

In traditional cloud computing platforms, Container Orchestrators are usually designed with heuristic policies without consideration of the diversity of workload scenarios and QoS requirements \cite{containerbasedsystem}. Their main drawbacks are listed as follows:

\begin{enumerate}
    \item Most of these policies are static heuristic methods configured offline according to certain workload scenarios at a limited scale. For instance, threshold-based autoscaling strategies can only be suitable for managing a set of pre-defined workloads \cite{costeffi,googletrace,stratus}. Such policies can not handle highly dynamic workloads where applications need to be scaled in/out at runtime according to specific behavior patterns.
    \item The performance of heuristic methods could dramatically degrade when the system scales up. For example, bin-packing algorithms, such as best fit or least fit algorithms, are widely utilised for solving task scheduling and resource allocation \cite{fitnessaware,containerpowereffi,video}. However, such algorithms could perform poorly with high task scheduling delays in large-scale compute clusters.
    \item Resource contention and performance interference between co-located applications are usually ignored. Co-located applications could compete for shared resources, which may cause application performance downgrade, extra maintenance costs, and service-level agreement (SLA) violations \cite{borg,alibabasemi,htas}. 
    \item The dependency structures between containerized application components are not considered during resource provisioning. Although some existing studies \cite{networkawareapp, powerawarescheduling, dynamicplacementmobile, multicompo} address this issue to a certain degree by leveraging machine learning (ML) methods, their ML models are only feasible for traditional cloud-based applications and relatively simple for containerized workload scenarios. For instance, containerized microservice-based applications are more lightweight and decentralized in nature, compared with traditional monolithic applications. There are internal connections between different microservice units within an application. Critical components in a microservice architecture are the dependencies of most other microservices and more likely to suffer from service level objectives (SLO) violations due to higher resource demands and communication costs \cite{sinan, firm}. These factors should all be taken into account during resource provisioning.
    \item Current container orchestration methodologies mostly emphasize the evaluation of infrastructure-level metrics, while application-level metrics and specific QoS requirements are not receiving sufficient consideration. For example, containerized workloads may be attached with stricter time constraints compared with traditional cloud workloads, such as task deployment delays, task completion time, and communication delays \cite{news,jobplacementdmlc}.
\end{enumerate}

With such fast-growing complexity of application management in cloud platforms, cloud service providers have a strong motivation to optimize their container orchestration policies by leveraging machine ML techniques \cite{towardmlcentric}. Through applying mathematical methods to training data, ML algorithms manage to build sophisticated analytic models that can precisely understand the behavior patterns in data flows or impacts of system operations. Thus, ML approaches could be adopted for modelling and prediction of both infrastructure-level or application-level metrics, such as application performance, workload characteristics, and system resource utilization. These insights could further assist the Container Orchestrators to improve the quality of resource provisioning decisions. On the other hand, ML algorithms could directly produce resource management decisions instead of heuristic methods, offering higher accuracy and lower time overhead in large-scale systems \cite{sarsa,negotiation,FScaler,migrationfog}. 

\subsection{Motivation of Research}
Machine learning-based container orchestration technologies have been leveraged in cloud computing environments for various purposes, such as resource efficiency, load balance, energy efficiency, and SLA assurance. Therefore, we aim to investigate them in depth in this paper:
\begin{enumerate}
 \item The machine learning-based container orchestration technologies have shown promise in application deployment and resource management in cloud computing environments. Hence, we aim to outline the evolution and principles of machine learning-based container orchestration technologies in existing studies.
 \item We recognize the need for a literature review to address the status of machine learning-based container orchestration researches in such fast-evolving and challenging scenarios. Furthermore, we investigate and classify the relevant papers by their key features. Our goal is to identify potential future directions and encourage more efforts in advanced research.
 \item  Although an innovative review is proposed on container orchestration in Reference \cite{containerbasedsystem}, the research on this field is growing continually by leveraging machine learning models. Therefore, there is a need for fresh reviews of machine learning-based container orchestration approaches to find out the advanced research challenges and potential future directions. 

\end{enumerate}

\subsection{Our Contributions}
The main contributions of this work are:
\begin{enumerate}
 \item We introduce a taxonomy of the most common ML algorithms used in the field of container orchestration.
 \item We present a comprehensive literature review of the state-of-the-art machine learning-based container orchestration approaches.
 \item We classify the reviewed orchestration methods by their key characteristics and conditions.
 \item We identify the future directions of machine learning-based container orchestration technologies.
\end{enumerate}

\subsection{Related Surveys}

\begin{table}[h]
\caption{A Comparison of Our Work with Existing Surveys Based on Key Parameters}
\label{table:ralted}
\begin{center} 
\tiny
\begin{tabular}{|c|c|c|c|c|c|c|c|c|c|}
\hline
\multirow{2}{*}{\textbf{Ref.}} & \multicolumn{2}{c|}{\textbf{\begin{tabular}[c]{@{}c@{}}Application Deployment \\ Unit\end{tabular}}} & \multicolumn{3}{c|}{\textbf{Infrastructure}}  & \multicolumn{4}{c|}{\textbf{Classification Schemes}}                                         \\ \cline{2-10} 
                               & \textbf{VM}                                   & \textbf{Container}                                   & \textbf{Cloud} & \textbf{Fog} & \textbf{Edge} & \textbf{\begin{tabular}[c]{@{}c@{}}Application \\ Architecture\end{tabular}} & \textbf{\begin{tabular}[c]{@{}c@{}}Behavior Modelling \\ and Prediction\end{tabular}} & \textbf{\begin{tabular}[c]{@{}c@{}}Resource \\ Provisioning\end{tabular}} & \textbf{\begin{tabular}[c]{@{}c@{}}ML-based \\ Orchestration Strategies\end{tabular}} \\ \hline
                               \cite{cloudresource} & \cmark & \cmark & \cmark & \xmark & \xmark & \xmark & \xmark & \cmark & \xmark \\ \hline
                               
                               \cite{brownout} & \cmark & \cmark & \cmark & \xmark & \xmark & \xmark & \xmark & \cmark & \xmark \\ \hline
                               
                               \cite{mlsurvey} & \cmark & \cmark & \cmark & \cmark & \cmark & \xmark & \cmark & \cmark & \cmark \\ \hline
                               
                               \cite{qosawarecloud} &  \cmark & \xmark & \cmark & \xmark & \xmark & \xmark & \xmark & \cmark & \xmark \\ \hline
                               
                               \cite{soatcontainertech} & \xmark & \cmark & \cmark & \cmark & \cmark & \xmark & \cmark & \cmark & \xmark \\ \hline
                               
                               \cite{cloudconainertechsoat} & \xmark & \cmark & \cmark & \cmark & \cmark & \cmark & \xmark & \cmark & \xmark \\ \hline
                               
                               \cite{containerbasedsystem} & \xmark & \cmark & \cmark & \xmark & \xmark & \xmark & \xmark & \cmark & \xmark \\ \hline
                               
                               This review & \xmark & \cmark & \cmark & \cmark & \cmark & \cmark & \cmark & \cmark & \cmark \\ \hline

\end{tabular}
\end{center}
\end{table}

As summarized in Table \ref{table:ralted}, some previous surveys have already explored the field of application management in cloud computing environments. Weerasiri et al. \cite{cloudresource} introduce a comprehensive classification framework for analysis and evaluation of cloud resource orchestration techniques. Xu and Buyya \cite{brownout} survey brownout technologies for adaptive application maintenance in cloud computing systems. Duc et al. \cite{mlsurvey} look into the research challenge of resource management and performance optimization under edge-cloud platforms, with emphasis on workload modelling and resource management through ML techniques. Singh and Chana \cite{qosawarecloud} depict a classification of QoS-aware autonomic resource provisioning under cloud computing environments through the methodical analysis of related research. However, these works are designed for managing general cloud workloads without adequate analysis of the key characteristics of containerized applications.

Furthermore, some recent studies also investigate different scopes of container orchestration techniques. Casalicchio and Iannucci \cite{soatcontainertech} conduct an extensive literature review of the state-of-the-art container technologies, focusing on three main issues, including performance evaluation, orchestration, and cyber-security. Pahl et al. \cite{cloudconainertechsoat} present a survey and taxonomy of cloud container technologies with a systematic classification of the existing researches. Rodriguez and Buyya  \cite{containerbasedsystem} propose a systematic review and taxonomy of the mainstream container orchestration systems by classifying their features, system architectures, and management strategies. Nonetheless, the research direction of machine learning-based orchestration for containerized applications has not been explored with systematic categories in any existing survey and taxonomy. 

Therefore, this paper extends the previous surveys by focusing on how machine learning algorithms could be applied to solve complex research problems from a container orchestration perspective, such as multi-dimensional workload characterization or autoscaling in hybrid clouds. We emphasis the diversity and complexity of orchestration schemes under specific application architectures and cloud infrastructures. Furthermore, it also specifies the extensive research questions and potential future directions of machine learning-based container orchestration techniques.

\subsection{Article Structure}

The rest of this paper is organized as follows: Section \ref{section:back} introduces the background of machine learning and container orchestration. Section \ref{section:tax} describes an overview of machine learning-based container orchestration technologies, followed by the categories and taxonomy of the existing approaches. A description of the reviewed approaches and their mappings to the categories are presented in Section \ref{section:review}. Section \ref{section:future} summarizes a discussion the future directions and open challenges. Finally, the conclusion of this work is given in Section \ref{section:conclude}.

\section{Background}
\label{section:back}

In this section, we present a comprehensive taxonomy of the existing ML models utilised in the area of container orchestration and a brief introduction of the high-level container orchestration framework.

\subsection{Machine Learning}

\begin{figure}[h]
 \centering
 \includegraphics[width=\linewidth]{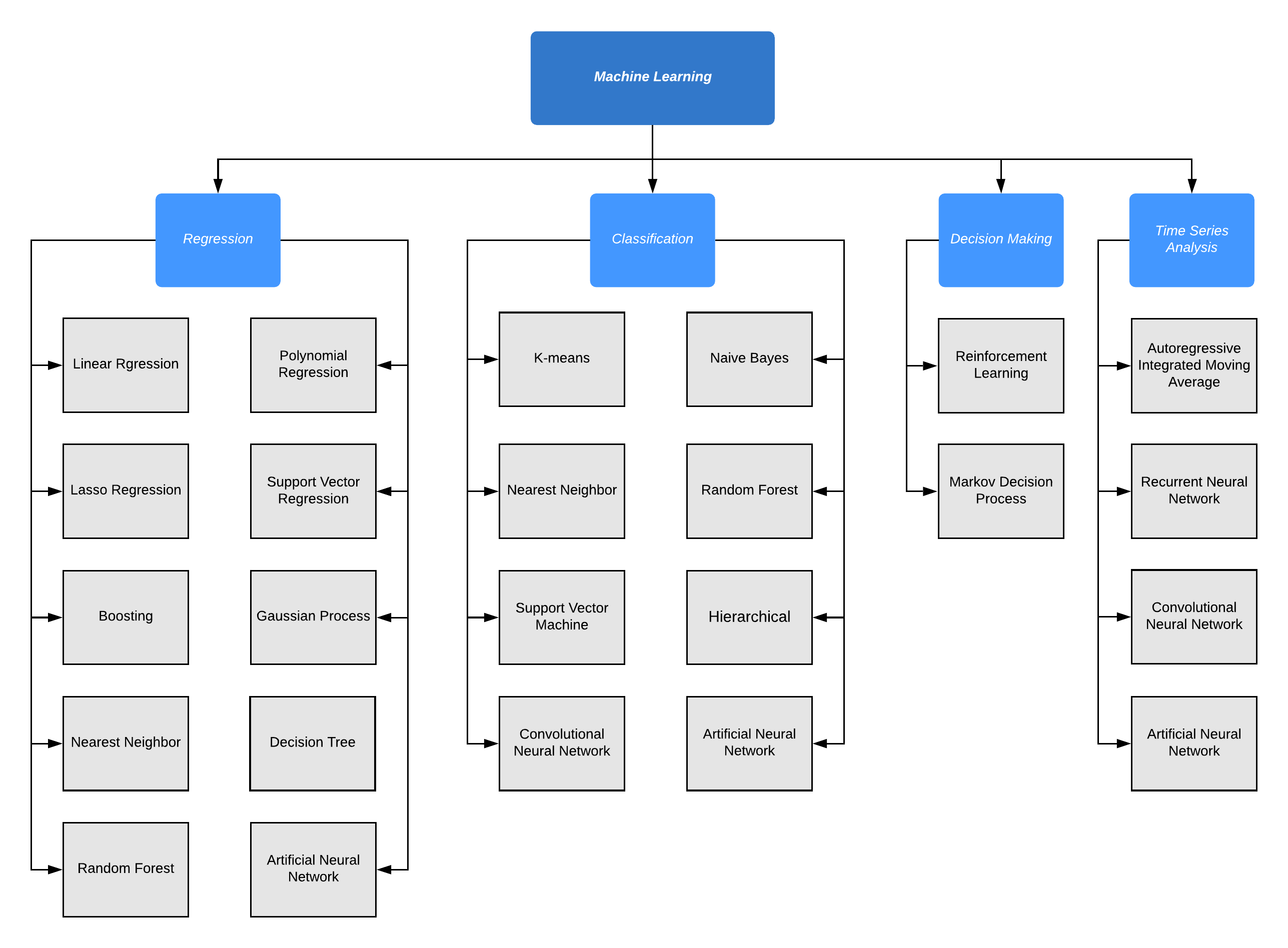}
 \caption{Machine Learning Taxonomy in the Context of Container Orchestration by Optimization Objectives}
 \Description{Machine Learning}
 \label{fig:ml}
\end{figure}

Machine learning is a general concept describing a set of computing algorithms that can learn from data and automatically build analytical models through data analysis \cite{mlreview}. One of its fundamental objectives is to build mathematical models that can emulate and predict the behavior patterns of various applications through training data. Machine learning has gained immense popularity through the past decades, widely accepted in many research areas such as image recognition, speech recognition, medical diagnose, and smart building \cite{mlstartbuilding,breastcancer}. With the continuously growing computational power and adoption of GPUs in warehouse-scale datacenters, the capability and data scale of machine learning technologies are still soaring \cite{deeplearning}.

As depicted in Fig. \ref{fig:ml}, we design a machine learning taxonomy through classifying some of the most popular ML models in the field of container orchestration by their optimization objectives:

\begin{enumerate}
    \item Regression algorithms predict a continuous output variable through analysis of the relationship between the output variable and one or multiple input variables. Popular regression algorithms (e.g., support vector regression, random forest, and polynomial regression) are usually used to explore and understand the relation between different performance metrics.
    
    \item Classification methods categorize training data into a series of classes. The use cases of classification mainly include anomaly detection and dependency analysis. For example, K-means is broadly adopted for the identification of abnormal system behaviors or components \cite{iotcontainer,containeranomalydetection}, while support vector machine (SVM) can be leveraged for decomposing the internal structure of containerized applications and identifying key application components \cite{firm}.
    
    \item Decision making models generate resolutions by simulating the decision making process and identifying the choices with the maximized cumulative rewards \cite{rlsurvey}. As the most common algorithms in this category, reinforcement learning (RL) models, including model-free (e.g., Q-Learning and Actor-Critic) and model-based RL, have been widely employed for decision making in resource provisioning.
    
    \item Time series analysis achieves pattern recognition of time series data and forecast of future time series values from past values. Ranging from autoregressive integrated moving average (ARIMA) to more advanced recurrent neural network (RNN) models, such algorithms are useful for behavior modelling of sequential data, including workload arrival rates or resource utilization.
\end{enumerate}

To be noted, some ML models can be utilised under multiple optimization scenarios. For example, artificial neural network (ANN) models can be applied for time series analysis of resource utilization or regression of application performance metrics \cite{sarsa,performancedocker}. 
\subsection{Container Orchestration}
\begin{figure}[h]
 \centering
 \includegraphics[width=\linewidth]{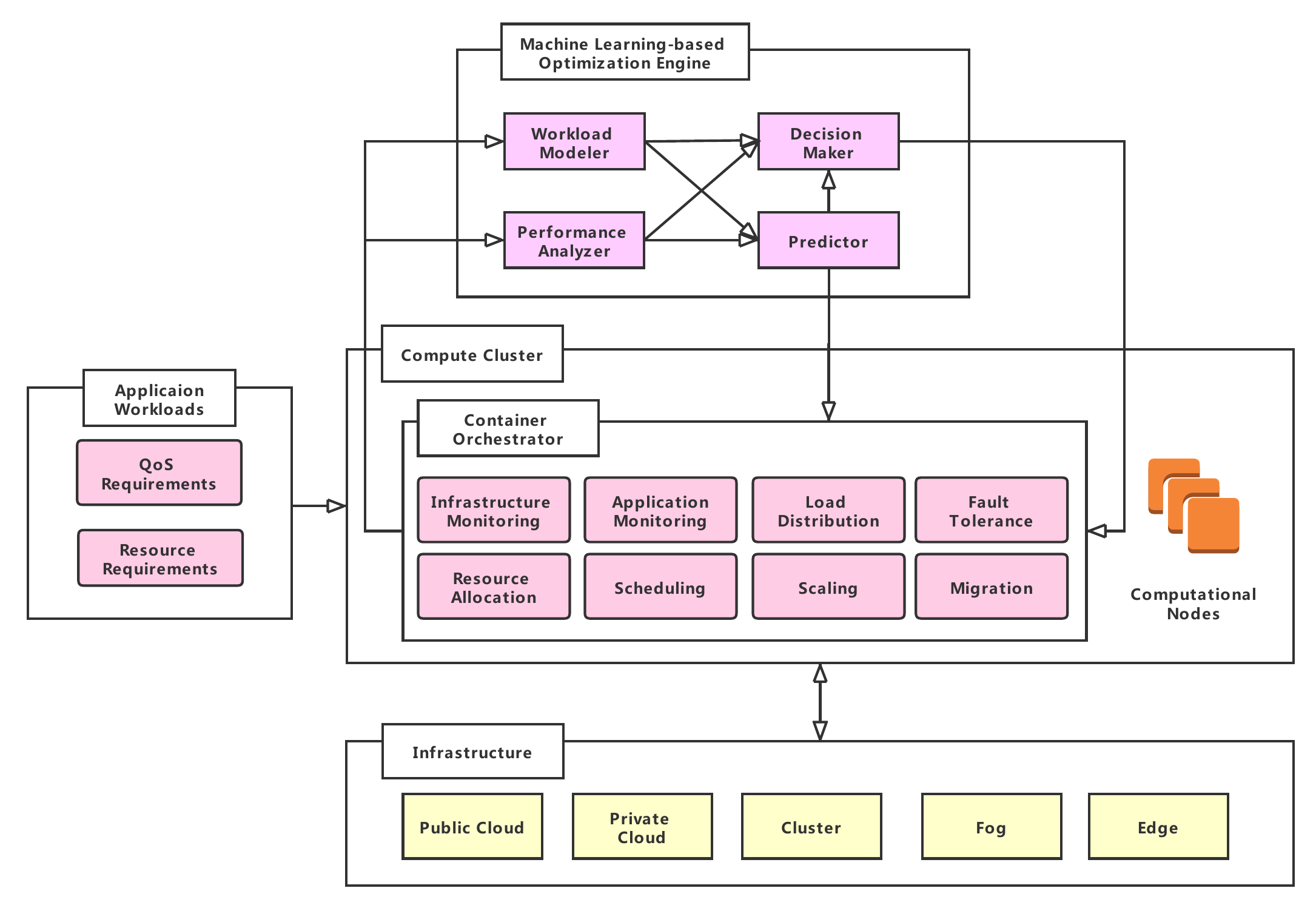}
 \caption{A High-level Machine Learning-based Container Orchestration Framework Reference Architecture}
 \Description{A High-level Machine Learning-based Container Orchestration Framework Reference Architecture}
 \label{fig:arc}
\end{figure}

Container orchestration empowers cloud service providers to decide how containerized applications are configured, deployed, and maintained under cloud computing environments \cite{soatcontainertech}. It is targeted at automatic application deployment and dynamic configuration adjustment at runtime for a diverse range of workloads. Fig. \ref{fig:arc} demonstrates a reference architecture of machine learning-based container orchestration frameworks, where the components included are common to most existing systems. A detailed description of each layer is given as below:

\subsubsection{Application Workloads} 

Workloads are generally defined as the requests submitted by users to an application or the Orchestrator. Workloads could be classified into many categories from different perspectives according to their unique features, such as long-running services or short-living batch jobs (classified by execution time); computation-intensive, data-intensive, or memory-sensitive workloads (classified by resource usage pattern) \cite{borg,alibabasemi,mixbatch}. Each workload is also defined with multi-dimensional resource requirements (e.g., CPU, memory, disk, network, etc.) and QoS requirements (e.g., time constraints, priorities, throughput, etc). The extremely dynamic and unpredictable feature of heterogeneous workloads greatly increases the complexity of orchestration mechanisms. 

\subsubsection{Compute Cluster} 

A compute cluster is a group of virtual or physical computational nodes that run in a shared network. As the core component in a containerized cluster, the Container Orchestrator is responsible for assigning containerized application components to worker nodes where containers are actually hosted and executed. Its major functionalities involve:

\begin{enumerate}
    \item Resource allocation assigns a specific amount of resources to a container. Such configurations and limitations are basic metrics in managing container placement and resource isolation control between containers.
    \item Scheduling defines the policy for the initial placement for one or a group of containers, by considering conditions such as resource constraints, communication costs, and QoS requirements. 
    \item Scaling is the resource configuration adjustment of containerized applications or computational nodes in response to any potential workload fluctuations. 
    \item Migration is designed as a complementary mechanism to prevent severe infrastructure-level resource overloading or resource contention between co-located applications by relocating one or a group of containers from one node to another.
    \item Infrastructure monitoring keeps the track of infrastructure-level metrics of the cluster, such as the number of nodes, node states, resource usage of each node, and network throughput. Such metrics are the critical information for evaluating the health conditions of the cluster and making precise decisions in the above resource source provisioning layers.
    \item Application monitoring does not only periodically check the application states but also records their resource consumption and performance metrics (e.g., response time, completion time, throughput, and error rates). This information serves as the crucial reference data in anomaly detection and SLA violation measurement.
    \item Load distribution, as the core mechanism for load balancing under containerized environments, distributes network traffic between containers evenly with policies such as Round-Robin \cite{roundrobin}. It plays an important role in improving system scalability, availability, and network performance. 
    \item Fault tolerance ensures the high availability of the system through replica control. Each container is maintained with a configurable number of replicas across multiple nodes in case of a single point of failure. It is also possible to have multiple Orchestrators to deal with unexpected node crashes or system overloading.

\end{enumerate}

\subsubsection{Infrastructure}

Thanks to the high portability, flexibility, and lightweight nature of containers, it allows containers to be deployed across a multitude of infrastructures, including private clouds, public clouds, fog, and edge devices. Similar to traditional server farms, private clouds provide exclusive resource sharing within a single organization based on its internal datacenter \cite{cloudcom}. By contrast, public cloud service providers support on-demand resource renting from their warehouse-scale cloud datacenters. 

Since applications and workloads are all deployed and processed at cloud datacenters in traditional cloud computing, this requires a massive volume of data transferred to cloud servers. As a consequence, it leads to significant propagation delays, communication costs, bandwidth and energy consumption. Through moving computation and storage facilities to the edge of a network, fog and edge infrastructures can achieve higher performance in a delay-sensitive, QoS-aware, and cost-saving manner \cite{edgesurvey, fogsurvey}. With the requests or data directly received from users, fog or edge nodes (e.g., IoT-enabled devices, routers, gateways, switches, and access points) can decide whether to host them locally or send them to the cloud. 

\subsubsection{Machine Learning-based Optimization Engine}
\label{subsubsection:mloe}
An ML-based optimization engine builds certain ML models for workload characterization and performance analysis, based on analysis of monitoring data and system logs received from the Orchestrator. Furthermore, it can produce future resource provisioning decisions relying on the generated behavior models and prediction results. The engine can be entirely integrated or independent from the Orchestrator. Its major components are listed as follows:

\begin{enumerate}
    \item Workload Modeller is designed for ML-based workload characterization through analyzing the input application workloads and identifying their key characteristics.
    \item Performance Analyzer generates application and system behavior models through applying ML algorithms to application and infrastructure-level monitoring data acquired from the Orchestrator.
    \item Predictor forms forecasts of workload volumes or application/system behaviors, relying on the models obtained from Workload Modeller and Performance Analyzer. The prediction results could be sent either to the Orchestrator or Decision Maker.
    \item Decision Maker combines the behavior models and prediction results received from the above components with certain ML-based optimization methods/schemes to further generate precise resource provisioning decisions that are fed back to the Orchestrator.
\end{enumerate}

These components are common to most existing ML-based approaches, however, not all of them have to be implemented to make the whole system functional. According to different design principles, the engine can alternatively produce multi-level prediction results to assist the original Orchestrator in making high-quality decisions in resource provisioning, or directly generate such decisions instead of the Orchestrator.

\section{Taxonomy of Machine Learning-based Container Orchestration Technologies}
\label{section:tax}

\begin{figure}[h]
 \centering
 \includegraphics[width=0.8\linewidth]{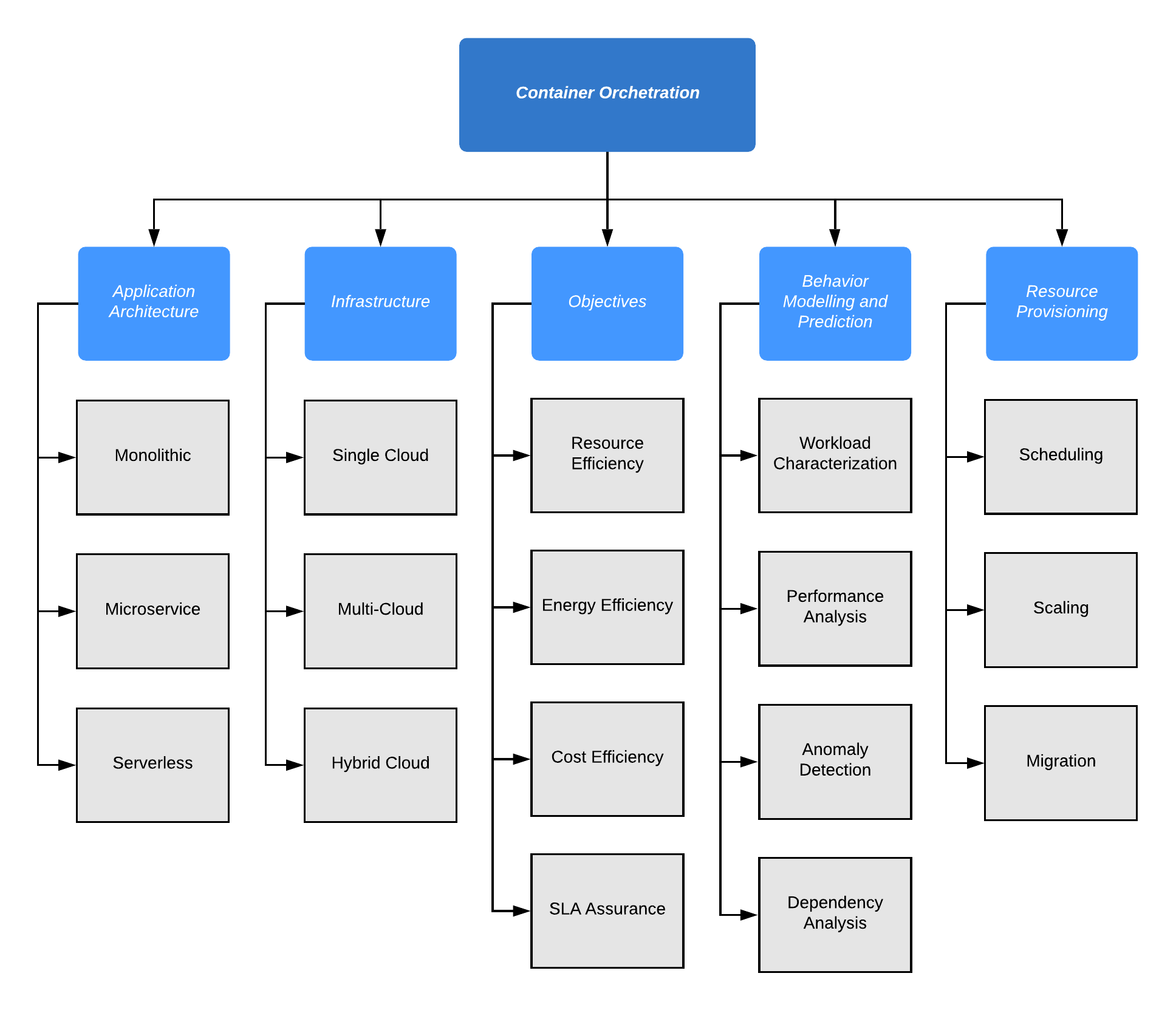}
 \caption{Machine Learning-based Container Orchestration Taxonomy}
 \Description{Container Orchestration}
 \label{fig:co}
\end{figure}

Fig. \ref{fig:co} presents a taxonomic classification of the literature reviewed in our work. Application Architecture describes the behaviors and internal structures of containerized application components that together perform as a whole unit. Infrastructure indicates the environments or platforms where the applications operate. Objectives are the improvements that the proposed ML-based approaches attempt to achieve. Behavior Modelling and Prediction leverage ML models for pattern recognition and simulation of system and application behaviors, as well as forecasting future trends according to previously collected data. Resource Provisioning specifies the ML-based or heuristic policies for resource management of containerized applications at different phases in a container lifecycle under various scenarios.

\begin{figure}[h]
 \centering
 \includegraphics[width=0.5\linewidth]{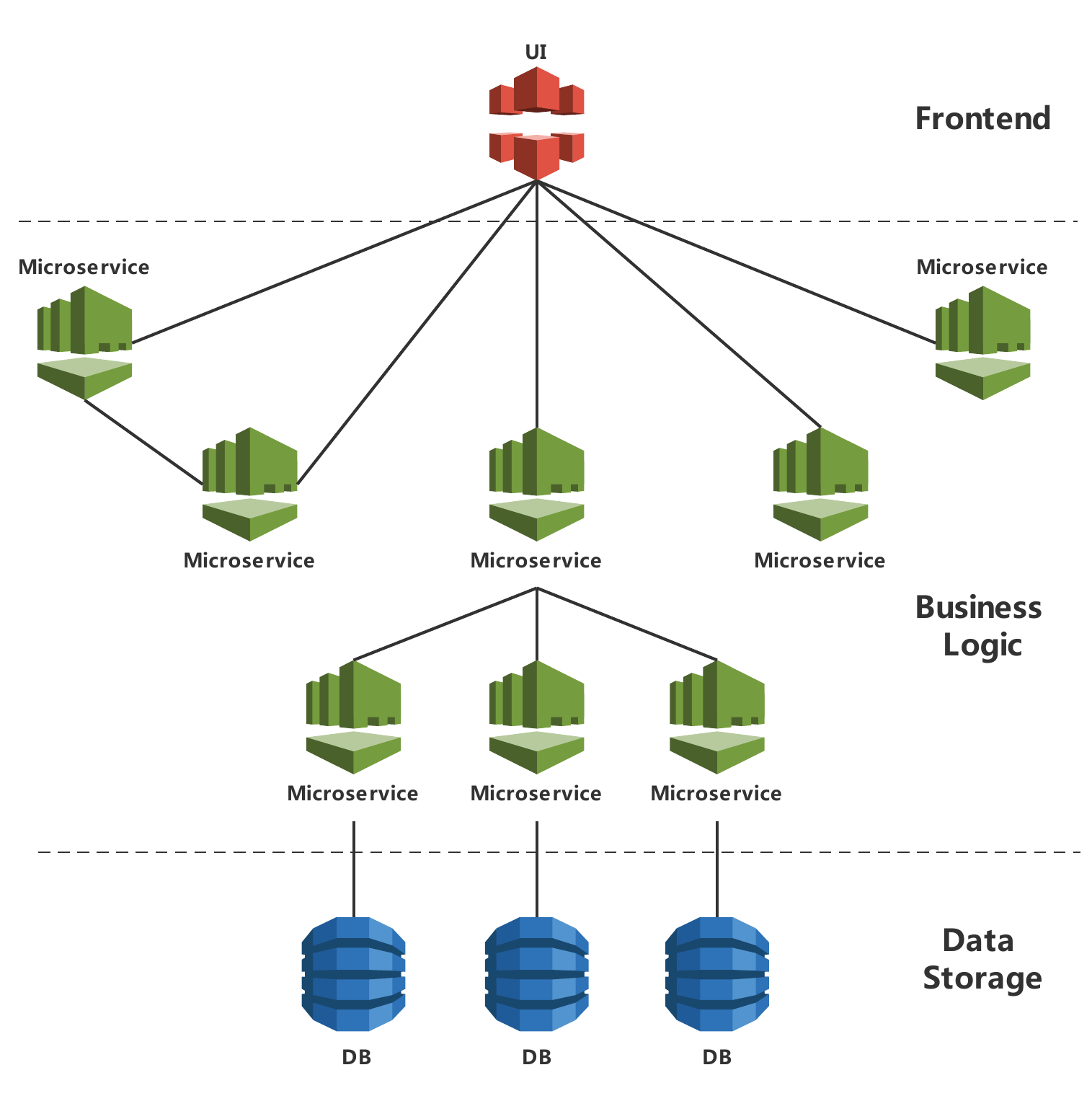}
 \caption{An Example of the Microservice Architecture}
 \Description{Microservice}
 \label{fig:micro}
\end{figure}

\subsection{Application Architecture}

This section presents the most common application architectures that define the composition of containerized applications and how they are deployed, executed, and maintained.

\subsubsection{Monolithic}

Monolithic applications follow an all-in-one architecture where all the functional modules are developed and configured into exactly one deployment unit, namely one container. Such applications could be initially easy to develop and maintain at a small scale. For example, Microsoft Azure \cite{azure} still supports automated single-container-based application deployment for enterprise solutions where the business logic is not feasible for building complex multi-component models. However, the consistent development and enrichment of monolithic applications would inevitably lead to incremental application sizes and complexity \cite{dsormono}. Consequently, the maintenance costs can dramatically grow in continuous deployment. Even modification of a single module requires retesting and redeployment of the whole application. Furthermore, scaling of monolithic applications means replication of the entire deployment unit containing all the modules \cite{webmigration}. In most scenarios, only a proportion of the modules need to be scaled due to resource shortage. Therefore, scaling the whole application would lead to poor overall resource efficiency and reliability. Thus, the monolithic architecture is only suitable for small-scale applications with simple internal structures.

\subsubsection{Microservice}

To address the problem of high development and maintenance costs caused by colossal application sizes, the microservice architecture (MSA) is proposed to split single-component applications into multiple loosely coupled and self-contained microservice components \cite{micromono}. An example of MSA is given in Fig. \ref{fig:micro}. Each microservice unit can be deployed and operate independently for different functionalities and business objectives. Furthermore, they can interact with each other and collaborate as a whole application through lightweight communication methods such as representational state transfer (REST). Through decomposing applications into a group of lightweight and independent microservice units, MSA has significantly reduced the costs and complexity of application development and deployment. Nonetheless, the growing number of components and dynamic inter-dependencies between microservices in MSA raises the problem of load distribution and resource management at the infrastructure level. A well-structured orchestration framework for MSA should be able to maintain multiple parts of an application with SLA assurance, granting more control over individual components.

\subsubsection{Serverless}

\begin{figure}[h]
 \centering
 \includegraphics[width=0.7\linewidth]{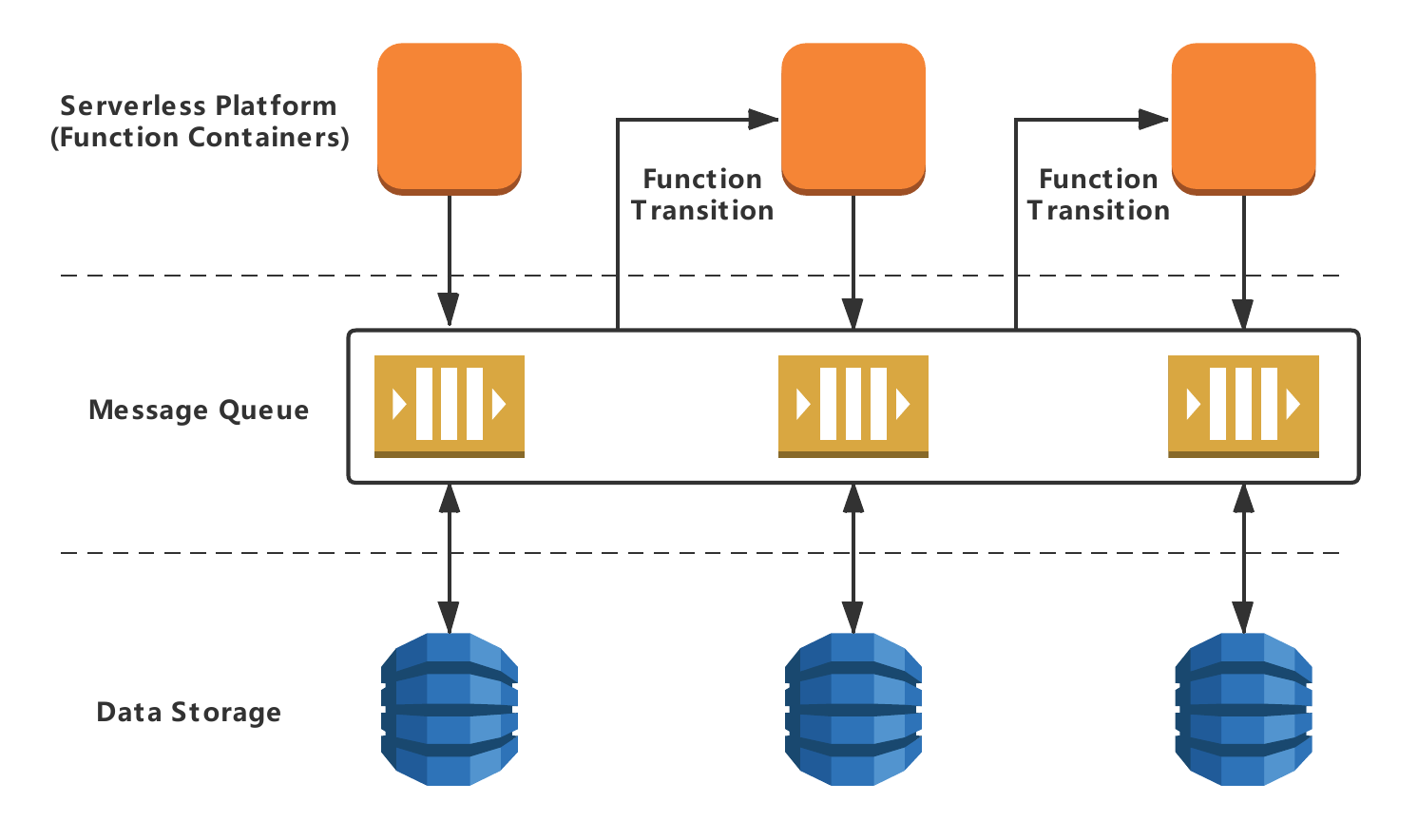}
 \caption{An Example of the Serverless Architecture}
 \Description{Serverless}
 \label{fig:serverless}
\end{figure}

The serverless architecture defines an event-driven application paradigm that undertakes stateless computational tasks, namely serverless functions. Designed to perform certain user-defined functionalities, functions are usually short pieces of code hosted in function containers with specific configurations and limited execution time. To ensure the successful completion and performance requirements of functions, serverless platforms are responsible for managing the function execution environments in terms of resource provisioning, energy consumption, SLA assurance, and security. On the one hand, this grants the platforms more control over application deployment and maintenance at the infrastructure level, while clients can focus on the development of application features and business logic. On the other hand, clients only have to pay for the resource consumption generated during function invocation. Overall, it provides clients with a cost-efficient pay-for-value billing standard as well as higher resource availability, scalability, and reliability. Thus, the serverless architecture has been adopted by most mainstream cloud service providers to automate workload-aware application deployment \cite{amazonlambda,serverlessreview}. 

As depicted in Fig. \ref{fig:serverless}, a typical serverless application is represented in the form of a function chain consisting of a workflow of individual functions. Within a function chain, interactions and transitions between functions are conducted through a centralized messaging service, such as a message queue or event bus \cite{serverlesscoldstart}. The orchestration problem under such context translates into managing the invocation of function chains in terms of function initialization, execution, transition, and data flow control. Several previous studies have proved that the significant time overhead of function initialization is currently the major performance bottleneck in function invocation \cite{catalyzer,firecracker}. Therefore,  
the current research focus of ML-based solutions in this field is on minimization of function invocation delays and resource wastage \cite{fifer,functionserverless,rlserverless}.

\subsection{Infrastructure}
A cloud infrastructure consists of a set of hardware resources and virtualization software to deliver virtualized resources to users for application deployment. In general, we have identified three types of cloud infrastructures in the context of container orchestration.

\begin{enumerate}
    \item A single cloud environment is built on resources from only one cloud service provider (either a private or public cloud) to host and serve all the applications.
    \item A multi-cloud environment includes multiple cloud services (e.g., private clouds, public clouds, or a mixture of both). As different cloud service providers may differ in many aspects, such as resource configurations, price, network latency, and geographic locations, this allows more choices for optimization of application deployment.
    \item A hybrid cloud environment is composed of a mixture of private clouds, public clouds, fog, or edge devices. It is not always efficient to deploy all the applications and data to public/private clouds, considering the data transmission time and network latency from end users to cloud servers. To solve this issue, the hybrid cloud enables applications and data to be deployed and processed at fog or edge devices that are close to end users.
\end{enumerate}

\subsection{Objectives}

In light of the diversity of application architectures and cloud infrastructures, many types of metrics have been considered as optimization objectives during the behavior modelling and resource provisioning of containerized applications. As presented in Fig. \ref{fig:co}, we group the existing objective metrics into four major categories. Since an orchestration solution usually needs to achieve multiple objectives according to specific user requirements, balancing the trade-off between different optimization objectives remains a key concern in automated application deployment and maintenance.

\begin{enumerate}

    \item \textbf{Resource Efficiency.} Infrastructure-level resource usage metrics are usually treated as key application performance indicators for energy and cost efficiency. They are the fundamental data source of most behavior modelling schemes, such as prediction of the resource demands of coming workloads, or discovery of the relationship between resource usage patterns and other performance metrics. Such insights could be further applied in decision making of resource provisioning to improve the overall resource efficiency and application performance.

    \item  \textbf{Energy Efficiency.} Under the continuously growing scale of cloud data centers, the tremendous electricity usage consumed by cloud infrastructures has emerged as a critical concern in the field of cloud computing \cite{energyeffi}. Therefore, various approaches have been proposed to minimize energy consumption and optimize energy efficiency \cite{iotcontainer,brownout,fifer,kuberknots}. As the overall electricity usage of a system is estimated as the summation of the consumption by each physical machine (PM) where its energy usage is directly related to its resource utilization, an essential way to control energy efficiency is to adjust and balance the resource utilization of physical machines during resource provisioning. 
    
    \item \textbf{Cost Efficiency.} Following the pay-as-you-go payment model of mainstream cloud service providers, market-based solutions regard cost efficiency as one of their principal targets \cite{costeffi,stratus,negotiation,selfadptive}. Through evaluation and selection of diverse cloud services according to their pricing models and computing capability, an optimized orchestration solution aims to minimize the overall financial costs while satisfying the QoS requirements defined by users. 
    
    \item \textbf{SLA Assurance.} Containerized applications are mostly configured with specific performance requirements, such as response time, initialization time, completion time, and throughput. These constraints are mostly expressed as SLA contracts, while their violations could lead to certain penalties. Because of the dynamic and unpredictable feature of cloud workloads, autoscaling is usually leveraged to automate application maintenance with SLA assurance \cite{rlserverless,htas,sarsa,microsacler,negotiation,Probabilistic,aiautoscale}, in response to the frequently changing workloads.
    
\end{enumerate}

\subsection{Behavior Modelling and Prediction}

 Behavior modelling is a fundamental step in understanding the overall application or system behavior patterns through analysis of application/infrastructure-level metrics. The variety of multi-layer metrics related to workloads, application performance, and system states significantly complicates the modelling process. Nonetheless, a well-structured behavior model that can produce precise prediction results is apparently useful for achieving certain optimization objectives during orchestration.

\begin{enumerate}
    \item Workload characterization captures the key features of application workloads. Because of the dynamic and decentralized nature of containerized applications like microservices, the received workloads may differ in many ways, such as task structures, resource demands, arrival rates, and distributed locations. These factors make it hard to define a robust method for characterization and categorization of all the different workloads within an orchestration system. However, the knowledge of workload behaviors is necessary to improve the quality of resource provisioning decisions by making precise resource assignments in response to any incoming workloads.
    \item Performance analysis discovers the relation among infrastructure (e.g., resource utilization and energy consumption) or application-level (e.g., response time, execution time, and throughput)
    metrics to depict the system states and application performance. These insights are important in managing the trade-off between different optimization objectives.
    \item Anomaly detection classifies and identifies abnormal system behaviors, including security threats, instance failure, workload spikes, performance downgrade, and resource overloading. Such anomalies could severely harm the system availability and reliability. Therefore, fast and accurate localization of their root causes could prevent SLA violations or system crashes.
    \item Dependency analysis looks into the graph-based internal dependencies between containerized application components. It helps to understand the workload distribution/pressure among application components and make precise resource configurations. Since the dependencies may be dynamically updated at runtime, it requires an incremental model that can consistently adjust itself and address the chain reactions of individual components to the overall application performance. 

\end{enumerate}

\subsection{Resource Provisioning}

\begin{figure}[h]
 \centering
 \includegraphics[width=0.7\linewidth]{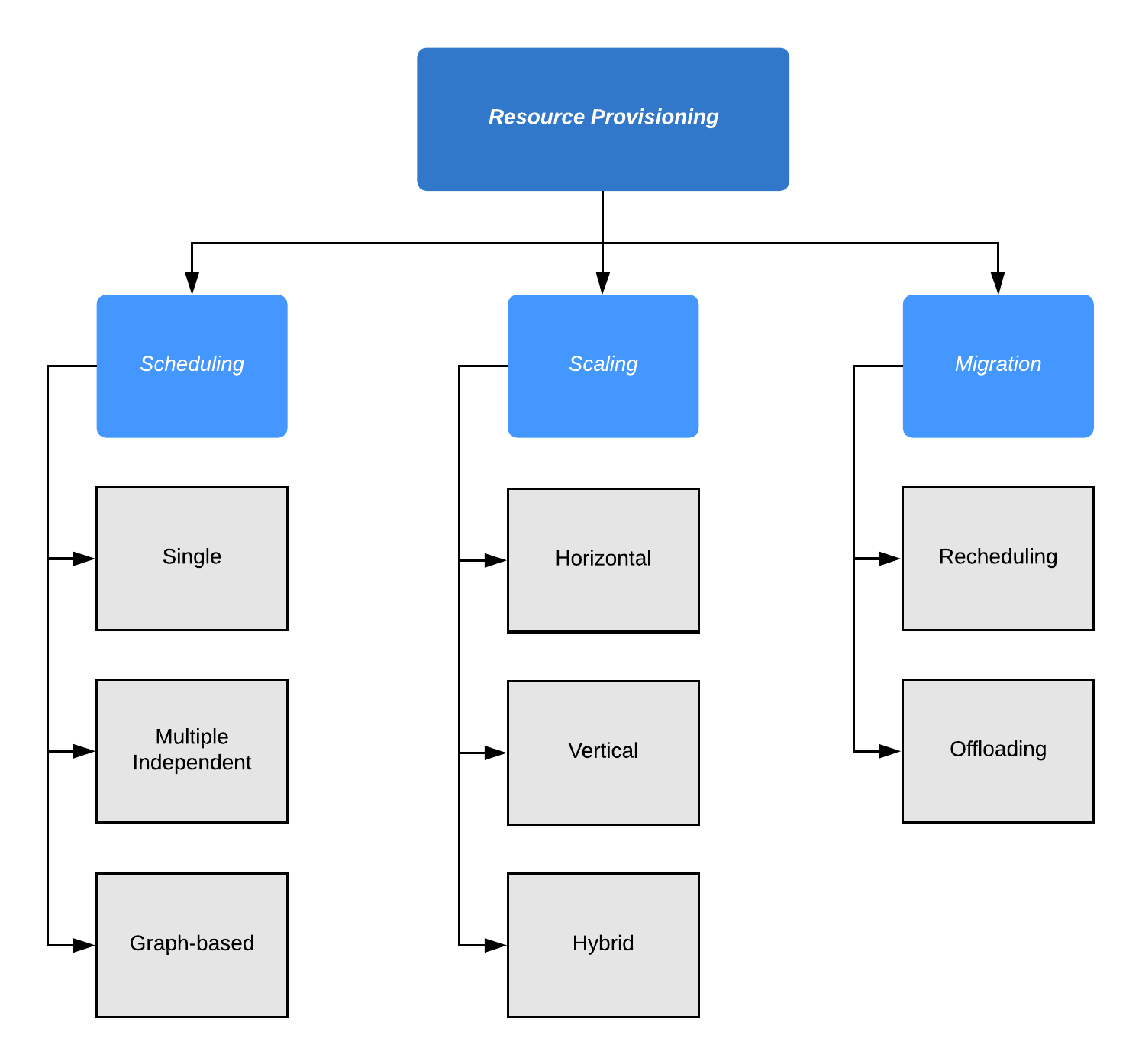}
 \caption{Resource Provisioning Taxonomy}
 \Description{RPT}
 \label{fig:rpt}
\end{figure}

Considering the various application architectures, infrastructures, and optimization objectives mentioned in the above sections, resource provisioning for containerized applications has become much more challenging. The diversity of cloud workloads, resource heterogeneity within hybrid cloud environments, and complex application internal structures should be all assessed during resource provisioning. Therefore, the state-of-the-art resource provisioning strategies are commonly relying on behavior modelling and prediction, to achieve higher accuracy and shorter computation delays. As addressed in Fig. \ref{fig:rpt}, resource provisioning operations can be classified into the following categories:

\begin{enumerate}
    \item Scheduling decides the initial placement of a containerized task unit, which could consist of a single task, a group of independent tasks, or a group of dependent tasks in graph-based structures. Due to the variety of application architectures and task structures, application deployment policies should consider a wide range of placement schemes. The quality of scheduling decisions has a direct impact on the overall application performance and resource efficiency \cite{microservicereview}.
    \item Scaling is the size adjustment of containerized applications or computational nodes in response to any potential workload fluctuations, which ensures the applications supported with enough resources to minimize SLA violations. Horizontal scaling adjusts the number of container replicas belonging to the applications or the number of nodes. By contrast, vertical scaling only updates the amount of resources assigned to existing containers or nodes. Moreover, hybrid scaling combines both horizontal and vertical scaling to produce an optimized solution.
    \item Migration is the relocation of one or a group of tasks from one node to another. When resource contention, overloading, or underloading occur between co-located applications due to poor scheduling decisions, rescheduling is triggered for load balancing through task migration within a single cloud. On the other hand, computation offloading manages the migration of computation-intensive tasks that are experiencing resource bottlenecks to devices with enough demanded resources across hybrid cloud environments. Targeted to improve the application performance, a refined offloading policy should be able to pick a suitable relocation destination that minimizes the migration costs in terms of execution time, bandwidth and energy consumption.
    
\end{enumerate}

\subsection{Evolution of Machine Learning-based Container Orchestration technologies}

\begin{figure}[h]
 \centering
 \includegraphics[width=\linewidth]{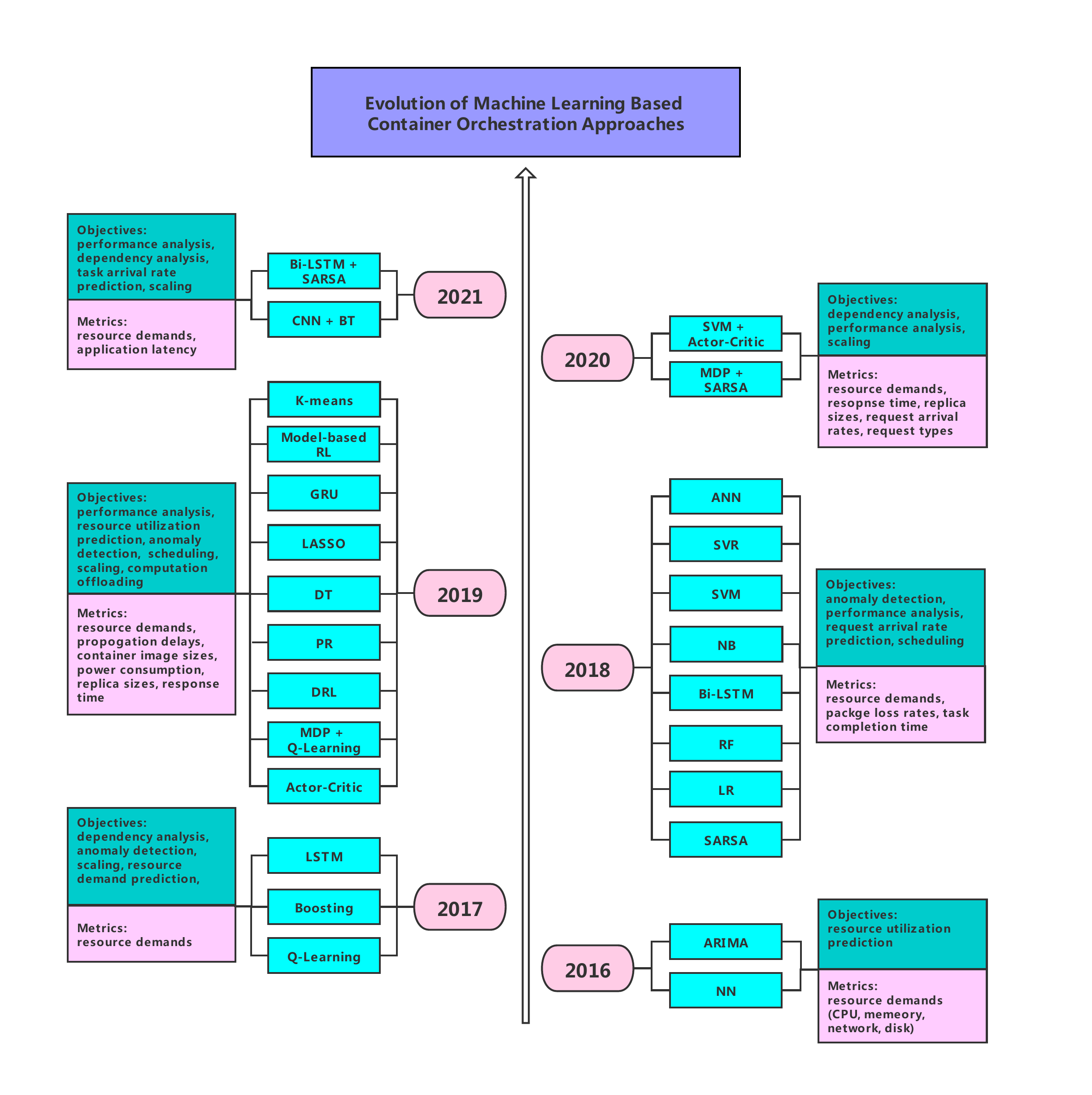}
 \caption{Evolution of Machine Learning-based Container Orchestration Technologies}
 \Description{Evolution of Machine Learning-based Container Orchestration Approaches}
 \label{fig:evolution}
\end{figure}

This section describes how ML-based methods are applied in the context of container orchestration. Fig. \ref{fig:evolution} demonstrates the evolution of ML-based models since 2016, with emphasis on their objectives and training metrics.

In 2016, the ARIMA \cite{CRUPA} and nearest neighbor (NN) \cite{video} algorithms were already leveraged for resource utilization prediction of containerized applications. At this stage, the application models were relatively simple, which only considered the time series pattern of infrastructure-level resource metrics.

In 2017, Shah et al. \cite{denpendencyana} first adopted the long short-term memory (LSTM) model for dependency analysis of microservices. Their model evaluated both the internal connections between microservice units and the time series pattern of resource metrics. Furthermore, anomaly detection was built on top of the LSTM model for the identification of abnormal behaviors in resource utilization or application performance. Besides, Cheng et al. \cite{highutiautoscaling} used Gradient Boosting Regression (GBR) for resource demand prediction in workload characterization. 

Xu et al. \cite{negotiation} leveraged a model-free RL method, namely Q-Learning, to produce vertical scaling plans. An optimal scaling decision was targeted to minimize resource wasting and computation costs under the assumption of SLA assurance.

In 2018, Tang et al. \cite{fisher} employed the bidirectional LSTM (Bi-LSTM) model for the prediction of workload arrival rates and application throughput. Their training module had demonstrated significant accuracy improvement over ARIMA and LSTM models in terms of time series prediction. Ye et al. \cite{performancedocker} applied a series of regression methods, including support vector regression (SVR), linear regression (LR), and ANN, to conduct performance analysis of relevant resource metrics. They attempted to evaluate the relationship between resource allocation and application performance. However, only single-component applications with limited performance benchmarks were considered within the scope of their work. 

Du et al. \cite{containeranomalydetection} designed an anomaly detection engine composed of k-nearest neighbors (KNN), SVM, Naive Bayes (NB), and random forest (RF), to classify and diagnose abnormal resource usage patterns in containerized applications. Orhean et al. \cite{news} utilised state–action–reward–state–action (SARSA), a model-free RL algorithm, to manage the graph-based task scheduling problem in directed acyclic graph (DAG) structures, aiming at minimizing the overall DAG execution time.

In 2019, Cheng et al. \cite{gruesprediction} proposed a hybrid gated recurrent unit (GRU) model to further reduce the computational costs and error rates of resource usage prediction of cloud workloads. Since LSTM models are relatively complex with high computational costs and data processing time, GRU is considered as an optimized model of LSTM \cite{gru}. An LSTM cell structure consists of three gates, including the input gate, the forget gate, and the output gate. GRU simplifies this structure and achieves higher computational efficiency by integrating the input gate and the forget gate into one update gate. 

Performance analysis of containerized applications was also explored in depth. Venkateswaran and Sarkar \cite{fitnessaware} leveraged the K-means clustering and polynomial regression (PR) to classify the multi-layer container execution structures under multi-cloud environments by their feasibility to the application performance requirements. According to workload arrival rates and resource metrics, Podolskiy et al. \cite{slocloudnative} applied Lasso regression (LASSO)  to forecast service level indicators (SLI) in terms of application response time and throughput. Dartois et al. \cite{ssdidperformance} used the decision tree (DT) regression algorithm to analyze the solid state drive (SSD) I/O performance under interference between applications. 

As for resource provisioning, deep reinforcement learning (DRL) was first applied in the context of task scheduling \cite{jobplacementdmlc,migrationfog}. Bao et al. \cite{jobplacementdmlc} designed a DRL framework for the placement of batch processing jobs, where an ANN model represented the mapping relationship between workload features, system states, and corresponding job placement decisions. The Actor-Critic RL algorithm was selected to train the ANN model and generate optimal scheduling decisions that minimized the performance interference between co-located batch jobs. Compared with traditional heuristic scheduling policies like bin packing, their solution demonstrated remarkable performance improvement regarding overall job execution time. Moreover, DRL was also employed to solve the problem of computation offloading under fog-cloud environments in Reference \cite{migrationfog}. On top of a Markov decision process (MDP) model that simulated the interactions of the offloading process at a large scale, the deep Q-Learning method optimized the migration decisions by minimization of the time overhead, energy usage, and computational costs. To explore the efficiency of hybrid scaling mechanisms, Rossi et al. \cite{hybridscaling,Geocontainer} leveraged model-based RL models to compose a mixture of horizontal and vertical scaling operations for monolithic applications, aiming at minimizing the resource usage, performance degradation, and adaption costs. 

In 2020, several RL-based scaling approaches were proposed in the form of hybrid ML models \cite{firm,FScaler,sarsa}. Qiu et al. \cite{firm} adopted the SVM model for dependency analysis of microservices and recognition of the key components that are highly likely to experience resource bottlenecks and performance downgrade. To prevent severe service level objectives (SLO) violations, the Actor-Critic method was utilised to generate the appropriate resource assignment decisions for these components through horizontal scaling. Besides, Sami et al. \cite{FScaler} combined MDP and SARSA models to build
a horizontal scaling solution for monolithic applications under fog-cloud environments. SARAS produced the optimized scaling decisions through model training relying on the MDP model that simulated the scaling scenarios with the fluctuating workloads and resource availability in fog taken into account.

In 2021, Zhang et al. \cite{sinan} proposed a novel approach composed of a convolutional neural network (CNN) and boosted trees (BT) for dependency and performance analysis of microservices. Their CNN model did not only analyze the inter-dependencies between microservice units for system complexity navigation, but also the time series metrics related to application performance. Furthermore, the BT model is responsible for the prediction of long-term QoS violations. To further improve the speed and efficiency of RL-based scaling approaches for microservices under hybrid cloud environments, Yan et al. \cite{HANSEL} developed a multi-agent parallel training module based on SARSA, supported by the microservice workload prediction results generated by Bi-LSTM.

Overall, diverse ML algorithms have been utilised in the context of container orchestration, ranging from workload modelling to decision making through RL. However, there are not many new ML models adopted in the area of container orchestration in recent years. To further improve prediction accuracy and computational efficiency, the emerging trend of hybrid ML-based solutions targets to combine multiple existing ML methods to form a complete orchestration pipeline, including multi-dimensional behavior modelling and resource provisioning. The evolution of ML models also contributes to the extension of various application architectures and cloud infrastructures. 

\section{State-of-the-Art in Machine Learning for Orchestration of Containers}
\label{section:review}

In this section, we introduce a literature review of machine learning-based container orchestration approaches. To stress the key features in the evaluated studies, we use the taxonomy in Section \ref{section:tax} to outline the key characteristics of the approaches designed for behavior modelling and prediction, as well as resource provisioning. The articles discussed in each class are key representative works selected from literature by several evaluation criteria, including robustness of application models, coverage of critical challenges in the field, accurate depiction of real-world orchestration scenarios, high scalability under complex cloud environments, novelty of orchestration approaches, refined combinations of ML-based models, and extensibility to hybrid clouds. 

\subsection{Behavior Modelling and Prediction}

This section presents the approaches related to behavior modelling and prediction. 

\subsubsection{Workload Characterization}
\label{subsubsection:wc}

\begin{table}[h]
\caption{Summary of Workload Characterization Approaches}
\label{table:wc}
\centering
\tiny
\begin{tabular}{|l|l|l|l|l|l|l|l|}
\hline
\textbf{Ref.}                                    & \textbf{\begin{tabular}[c]{@{}l@{}}Application \\ Architecture\end{tabular}} & \textbf{Methods}                                         & \textbf{Type}                                                   & \textbf{Infrastructure} & \textbf{Objectives}                                                                                         & \textbf{Advantages}                                                                        & \textbf{Limitations}                                                                                                                      \\ \hline
\cite{fifer}                    & Serverless                                                                   & LSTM                                                     & \begin{tabular}[c]{@{}l@{}}Time series \\ analysis\end{tabular} & Single cloud            & \begin{tabular}[c]{@{}l@{}}Request arrival rate \\ prediction\end{tabular}                                  & High prediction accuracy                                                                   & Simplicity of application models                                                                                                          \\ \hline
\cite{htas}                     & Monolithic                                                                   & K-means++                                                & Classification                                                  & Single cloud            & \begin{tabular}[c]{@{}l@{}}Resource demand \\ prediction\end{tabular}                                       & High scalability                                                                           & \begin{tabular}[c]{@{}l@{}}Limited accuracy under high load \\ variance\end{tabular}                                                      \\ \hline
\cite{sarsa}                    & Monolithic                                                                   & ARIMA                                                    & \begin{tabular}[c]{@{}l@{}}Time series \\ analysis\end{tabular} & Single cloud            & \begin{tabular}[c]{@{}l@{}}Request arrival rate\\  prediction\end{tabular}                                  & Capability of large data scales                                                             & Inaccuracy under trend turning                                                                                                            \\ \hline
\cite{hybridarima}              & Monolithic                                                                   & ARIMA-TES                                                & \begin{tabular}[c]{@{}l@{}}Time series \\ analysis\end{tabular} & Single cloud            & \begin{tabular}[c]{@{}l@{}}Multi-dimensional \\ workload prediction\end{tabular}                            & \begin{tabular}[c]{@{}l@{}}High robustness, accuracy,\\ and anti-interference \end{tabular} & Higher time overhead                                                                                                                      \\ \hline
\cite{hybridworkloadprediction} & Monolithic                                                                   & \begin{tabular}[c]{@{}l@{}}LSTM, \\ Bi-LSTM\end{tabular} & \begin{tabular}[c]{@{}l@{}}Time series \\ analysis\end{tabular} & Single cloud            & \begin{tabular}[c]{@{}l@{}}Resource demand \\ prediction\end{tabular}                                       & High prediction accuracy                                                                   & \begin{tabular}[c]{@{}l@{}}Lack of consideration for anomaly \\ detection\end{tabular}                                                    \\ \hline
\cite{gruesprediction}          & Monolithic                                                                   & GRU-ES                                                   & \begin{tabular}[c]{@{}l@{}}Time series \\ analysis\end{tabular} & Single cloud            & \begin{tabular}[c]{@{}l@{}}Resource demand \\ prediction\end{tabular}                                       & Low error rates                                                                            & \begin{tabular}[c]{@{}l@{}}Ignorance of potential resource\\  allocation strategies based on \\ the prediction model\end{tabular}         \\ \hline
\cite{video}                    & Microservice                                                                 & TSNNR                                                    & Regression                                                      & Single cloud            & \begin{tabular}[c]{@{}l@{}}Resource demand \\ prediction\end{tabular}                                       & Improved prediction accuracy                                                               & Limited workload scenarios                                                                                                                \\ \hline
\cite{highutiautoscaling}       & Microservice                                                                 & GBR                                                      & Regression                                                      & Single cloud            & \begin{tabular}[c]{@{}l@{}}Resource demand \\ prediction\end{tabular}                                       & High prediction accuracy                                                                   & Limited workload scenarios                                                                                                                \\ \hline
\cite{mlbasedautoscaling}       & Microservice                                                                 & LSTM                                                     & \begin{tabular}[c]{@{}l@{}}Time series \\ analysis\end{tabular} & Single cloud            & \begin{tabular}[c]{@{}l@{}}Request arrival rate\\  prediction\end{tabular}                                  & Low time overhead                                                                          & High computational expense                                                                                                                \\ \hline
\cite{gruprediction}            & Microservice                                                                 & IGRU-SD                                                  & \begin{tabular}[c]{@{}l@{}}Time series \\ analysis\end{tabular} & Hybrid cloud            & \begin{tabular}[c]{@{}l@{}}Resource demand \\ prediction\end{tabular}                                       & Low error rates                                                                            & \begin{tabular}[c]{@{}l@{}}Unclear demonstration of the \\ relationship between resource\\  allocation and energy efficiency\end{tabular} \\ \hline
\cite{aiautoscale}              & Microservice                                                                 & \begin{tabular}[c]{@{}l@{}}AR, LSTM, \\ HTM\end{tabular} & \begin{tabular}[c]{@{}l@{}}Time series \\ analysis\end{tabular} & Single cloud            & \begin{tabular}[c]{@{}l@{}}Request arrival rate\\  prediction\end{tabular}                                  & Multi-model optimization                                                                   & High computational costs                                                                                                                  \\ \hline
\cite{fisher}                   & Microservice                                                                 & Bi-LSTM                                                  & \begin{tabular}[c]{@{}l@{}}Time series \\ analysis\end{tabular} & Single cloud            & \begin{tabular}[c]{@{}l@{}}Prediction of request \\ arrival rate and \\ application throughput\end{tabular} & Improved prediction accuracy                                                               & Implicit time overhead analysis                                                                                                           \\ \hline
\cite{HANSEL}                   & Microservice                                                                 & Bi-LSTM                                                  & \begin{tabular}[c]{@{}l@{}}Time series \\ analysis\end{tabular} & Hybrid cloud            & \begin{tabular}[c]{@{}l@{}}Task arrival rate \\ prediction\end{tabular}                                     & Performance improvement                                                                    & Inaccuracy in long-term forecasts                                                                                                         \\ \hline
\end{tabular}
\end{table}

The knowledge of workload characteristics and behavior patterns offers important reference data for the estimation of resource provisioning and load distribution. Table \ref{table:wc} shows the existing studies that leverage ML techniques in workload characterization. 

Many previous studies have tried to model the time series pattern of request arrival rates in containerized applications through various algorithms. ARIMA, as a classic algorithm for analyzing time series data, was utilised in Reference \cite{sarsa}. Compared with other linear models such as autoregressive (AR), moving average (MA), autoregressive moving average (ARMA), and exponentially weighted moving average (EWMA), ARIMA enjoys high accuracy even under unstable time series. As the workload scenario of ARIMA is limited to linear models, it is usually referenced as a baseline approach by many studies included in our literature review.

To further speed up the data training process, LSTM models are adopted in References \cite{fifer,mlbasedautoscaling,aiautoscale} to predict the request arrival rates under large dynamic variations and avoid unnecessary resource provisioning operations. Unlike general feedforward neural networks, LSTM has feedback connections. It produces prediction results by analyzing the whole data sequences and are more accurate at identifying new patterns. However, LSTM models only train the data in one direction. Bi-LSTM models can process the data sequence from both forward and backward directions. Therefore, Bi-LSTM models are proposed in References \cite{fisher,HANSEL} to capture more key metrics and improve the prediction accuracy. 

Another direction in ML-based workload characterization is resource demand modelling and prediction. Zhong et al. \cite{htas} leverage the K-means++ algorithm for task classification and identification based on the resource usage (e.g., CPU and memory) patterns of different workloads. Zhang et al. \cite{video} introduce the Time Series Nearest Neighbor Regression (TSNNR) algorithm for prediction of future workload resource requirements by matching the recent time series data trend to similar historical data. To enhance the ARIMA model in analysis of nonlinear relationship and trend turning points within data sequences,  Xie et al. \cite{hybridarima} apply ARIMA with triple exponential smoothing (ARIMA-TES) in the prediction of container workload resource usage. 

Besides, A hybrid association learning architecture is designed in Reference \cite{hybridworkloadprediction} through combining the LSTM and Bi-LSTM models. A multi-layer structure is built to find the inter-dependencies and relationship between various resource metrics, which are generally classified into three distinct groups, including CPU, memory, and I/O. To further reduce the error rates of resource usage prediction and data training time, GRU-based models are utilised due to their high computational efficiency and prediction accuracy. Lu et al. \cite{gruprediction} develop a hybrid prediction framework consisting of a GRU and a straggler detection model (IGRU-SD) to predict the periodical resource demand patterns of cloud workloads on a long-term basis. Likewise, Cheng et al. \cite{gruesprediction} propose a GRU-ES model where the exponential smoothing method is used to update the resource usage prediction results generated by GRU and reduce prediction errors.

In summary, the majority of the reviewed literature in workload characterization is focusing on the analysis and prediction of request arrival rates and resource usage patterns through time series analysis or regression models.

\subsubsection{Performance Analysis}
\label{subsubsection:bmp}

\begin{table}[h]
\caption{Summary of Performance Analysis Approaches}
\label{table:bmp}
\centering
\tiny
\begin{tabular}{|l|l|l|l|l|l|l|l|}
\hline
\textbf{Ref.}                               & \textbf{\begin{tabular}[c]{@{}l@{}}Application \\ Architecture\end{tabular}} & \textbf{Methods}                                                  & \textbf{Type}                                                                  & \textbf{Infrastructure} & \textbf{Objectives}                                                                                                   & \textbf{Advantages}                                                                        & \textbf{Limitations}                                                                          \\ \hline
\cite{functionserverless}  & Serverless                                                                   & LSTM, LR                                                          & \begin{tabular}[c]{@{}l@{}}Time series \\ analysis, \\ regression\end{tabular} & Single cloud            & \begin{tabular}[c]{@{}l@{}}Prediction of function \\ invoking time\end{tabular}                                       & \begin{tabular}[c]{@{}l@{}}Online prediction of \\ function chains\end{tabular}            & \begin{tabular}[c]{@{}l@{}}Additional resource consumption \\ for model training\end{tabular} \\ \hline
\cite{edgecloudserverless} & Serverless                                                                   & GBR                                                               & Regression                                                                     & Hybrid cloud            & \begin{tabular}[c]{@{}l@{}}Prediction of costs and \\ end-to-end latency\end{tabular}                                 & High prediction accuracy                                                                   & High computational expenses                                                                   \\ \hline
\cite{cose}                & Serverless                                                                   & BO                                                                & Regression                                                                     & Hybrid cloud            & \begin{tabular}[c]{@{}l@{}}Prediction of costs and \\ execution time based on \\ function configurations\end{tabular} & High prediction accuracy                                                                   & High computational complexity                                                                 \\ \hline
\cite{fitnessaware}        & Monolithic                                                                   & K-means, PR                                                       & \begin{tabular}[c]{@{}l@{}}Classification, \\ regression\end{tabular}          & Hybrid cloud            & \begin{tabular}[c]{@{}l@{}}Classification of cloud \\ service providers and \\ container clusters\end{tabular}        & \begin{tabular}[c]{@{}l@{}}High prediction accuracy \\ and effectiveness\end{tabular}      & Simplicity of application models                                                              \\ \hline
\cite{sarsa}               & Monolithic                                                                   & ANN                                                               & \begin{tabular}[c]{@{}l@{}}Time series \\ analysis\end{tabular}                & Single cloud            & \begin{tabular}[c]{@{}l@{}}Resource utilization and \\ response time prediction\end{tabular}                          & Incremental modelling                                                                      & Long model training time                                                                      \\ \hline
\cite{kuberknots}          & Monolithic                                                                   & ARIMA                                                             & \begin{tabular}[c]{@{}l@{}}Time series \\ analysis\end{tabular}                & Single cloud            & \begin{tabular}[c]{@{}l@{}}GPU resource utilization \\ prediction\end{tabular}                                        & \begin{tabular}[c]{@{}l@{}}Discovery of the peak of \\ resource consumption\end{tabular}   & \begin{tabular}[c]{@{}l@{}}Implicit accuracy and time \\ overhead evaluation\end{tabular}     \\ \hline
\cite{performancedocker}   & Monolithic                                                                   & \begin{tabular}[c]{@{}l@{}}SVR, LR, \\ ANN\end{tabular}           & Regression                                                                     & Single cloud            & \begin{tabular}[c]{@{}l@{}}Prediction of application \\ performance based on \\ resource metrics\end{tabular}         & Low error rates                                                                            & High computational expenses                                                                   \\ \hline
\cite{miras}               & Microservice                                                                 & ANN                                                               & Regression                                                                     & Single cloud            & \begin{tabular}[c]{@{}l@{}}Performance modelling\\ of microservice workflow \\ systems\end{tabular}                   & \begin{tabular}[c]{@{}l@{}}Sample complexity \\ reduction\end{tabular}                     & \begin{tabular}[c]{@{}l@{}}Randomness due to boundary \\ effects\end{tabular}                 \\ \hline
\cite{loadbalancemicro}    & Microservice                                                                 & SVR, ANN                                                          & Regression                                                                     & Multi-cloud             & \begin{tabular}[c]{@{}l@{}}Performance modelling \\ of microservice response \\ time\end{tabular}                     & \begin{tabular}[c]{@{}l@{}}Time saving for route \\ searching\end{tabular}                 & \begin{tabular}[c]{@{}l@{}}Dependency on offline training\\  of historical data\end{tabular}  \\ \hline
\cite{containeroffloading} & Microservice                                                                 & \begin{tabular}[c]{@{}l@{}}LR, PR, RF, \\ SVR\end{tabular}        & Regression                                                                     & Hybrid cloud            & \begin{tabular}[c]{@{}l@{}}Prediction of offloading \\ execution time\end{tabular}                                    & High prediction accuracy                                                                   & High computational costs                                                                      \\ \hline
\cite{Probabilistic}       & Microservice                                                                 & GP                                                                & Regression                                                                     & Single cloud            & \begin{tabular}[c]{@{}l@{}}Prediction of end-to-end \\ latency\end{tabular}                                           & High prediction accuracy                                                                   & High computational complexity                                                                 \\ \hline
\cite{slocloudnative}      & Microservice                                                                 & \begin{tabular}[c]{@{}l@{}}LR, RF, \\ LASSO\end{tabular}          & Regression                                                                     & Single cloud            & Prediction of SLI                                                                                                     & \begin{tabular}[c]{@{}l@{}}High prediction accuracy \\ by removing anomalies\end{tabular}  & High computational complexity                                                                 \\ \hline
\cite{sinan}               & Microservice                                                                 & CNN, BT                                                           & \begin{tabular}[c]{@{}l@{}}Time series \\ analysis, \\ regression\end{tabular} & Single cloud            & \begin{tabular}[c]{@{}l@{}}End-to-end latency and \\ QoS violations prediction\end{tabular}                           & \begin{tabular}[c]{@{}l@{}}Online prediction with high \\ resource efficiency\end{tabular} & Overfitting and misprediction                                                                 \\ \hline
\cite{ssdidperformance}    & Microservice                                                                 & \begin{tabular}[c]{@{}l@{}}DT, MARS, \\ boosting, RF\end{tabular} & Regression                                                                     & Single cloud            & \begin{tabular}[c]{@{}l@{}}Modelling and prediction \\ of SSD I/O performance\end{tabular}                            & Short data training time                                                                   & Simplicity of application models                                                              \\ \hline
\end{tabular}
\end{table}

Performance analysis captures the key infrastructure and application-level metrics for evaluation of the overall system status or application performance. A summary of the ML-based performance analysis approaches is given in Table \ref{table:bmp}.

Das et al. \cite{edgecloudserverless} design a performance modeller based on GBR for predicting the costs and end-to-end latency of input serverless functions based on their configurations under edge-cloud environments. Such metrics are the key factors in the estimation of the efficiency of function scheduling decisions. Similarly, Akhtar et al. \cite{cose} leverage the Bayesian Optimization (BO) function to achieve the same purpose. The prediction results will be used to estimate the optimal function configurations that meet the time constraints in function deployment with the lowest costs. 

To estimate the cold start latency in serverless computing platforms, Xu et al. \cite{functionserverless} propose a two-phase approach. LSTM is used in the first phase to predict the invoking time of the first function in a function chain, while the rest of the functions is processed through LR. This two-phase approach achieves a significant reduction of prediction error rates and resource wasting in the container pool.

Venkateswaran and Sarkar\cite{fitnessaware} try to classify the cloud service providers and container cluster configurations under multi-cloud environments through a two-level approach.  K-means is employed in the first level to precisely classify the comparable container cluster compositions by their performance data, while PR is applied in the second level for analyzing the relationship between container strength and container system performance. 

Some research works have investigated the issue of infrastructure-level resource utilization prediction \cite{sarsa,kuberknots,ssdidperformance}. Zhange et al. \cite{sarsa} leverage the ANN model to predict the CPU utilization and request response time by modelling a series of metrics, including CPU and memory usage, response time, and the request arrival rates mentioned in Section \ref{subsubsection:wc}. Besides, Dartois et al. \cite{ssdidperformance} look into the research topic of SSD I/O performance modelling and interference prevention with a series of regression techniques, including boosting, RF, DT, and Multivariate adaptive regression splines (MARS). As SSDs are prevalently used by large-scale datacenter for data storage due to their high performance and energy efficiency, their internal mechanisms could directly impact application-level behaviors and cause potential SLO violations.  

On the other hand, some previous studies focus on behavior modelling of application-level metrics \cite{slocloudnative,sinan,Probabilistic,miras}. RScale \cite{Probabilistic} is implemented as a robust scaling system leveraging Gaussian process (GP) regression for investigation of the interconnections between end-to-end tail latency of microservice workloads and internal performance dependencies. Likewise, Sinan \cite{sinan}, as an ML-based and QoS-aware cluster manager for containerized microservices, combines the CNN and BT model for the prediction of end-to-end latency and QoS violations. Taking both the microservice inter-dependencies and the time series pattern of application-level metrics into account, Sinan evaluates the efficiency of short-term resource allocation decisions as well as long-term application QoS performance.   

Overall, most studies in this section are concerned with prediction of time constraints or resource usage patterns through regression or time series analysis techniques.

\subsubsection{Anomaly Detection}
\label{subsubsection:ad}

\begin{table}[h]
\caption{Summary of Anomaly Detection Approaches}
\label{table:ad}
\centering
\tiny
\begin{tabular}{|l|l|l|l|l|l|l|l|}
\hline
\textbf{Ref.}                                     & \textbf{\begin{tabular}[c]{@{}l@{}}Application \\ Architecture\end{tabular}} & \textbf{Methods}                                                                & \textbf{Type}                                                   & \textbf{Infrastructure} & \textbf{Objectives}                                                                                                           & \textbf{Advantages}                                                     & \textbf{Limitations}                                                                              \\ \hline
\cite{iotcontainer}              & Monolithic                                                                   & \begin{tabular}[c]{@{}l@{}}K-means, ensemble, \\ hierarchical\end{tabular}      & Classification                                                  & Hybrid cloud            & \begin{tabular}[c]{@{}l@{}}Identification of overloaded \\ or underloaded nodes\end{tabular}                                  & \begin{tabular}[c]{@{}l@{}}Short data training \\ time\end{tabular}     & \begin{tabular}[c]{@{}l@{}}Limited workload scenarios, \\ high space complexity\end{tabular}      \\ \hline
\cite{containerdetection}        & Monolithic                                                                   & \begin{tabular}[c]{@{}l@{}}K-means, KNN, \\ self-organizing map\end{tabular} & Classification                                                  & Single cloud            & \begin{tabular}[c]{@{}l@{}}Container vulnerability \\ detection\end{tabular}                                                  & \begin{tabular}[c]{@{}l@{}}High detection \\ accuracy\end{tabular}      & Insufficient anomaly cases                                                                                \\ \hline
\cite{containeranomalydetection} & Microservice                                                                 & KNN, SVM, NB, RF                                                                & Classification                                                  & Single cloud            & \begin{tabular}[c]{@{}l@{}}Anomaly detection of \\ microservices according to \\ real-time performance metrics\end{tabular}   & \begin{tabular}[c]{@{}l@{}}Various monitoring \\ metrics\end{tabular}   & \begin{tabular}[c]{@{}l@{}}Insufficient evaluation of \\ application-level anomalies\end{tabular} \\ \hline
\cite{denpendencyana}            & Microservice                                                                 & LSTM                                                                            & \begin{tabular}[c]{@{}l@{}}Time series \\ analysis\end{tabular} & Single cloud            & \begin{tabular}[c]{@{}l@{}}Anomaly detection and \\ prediction on application \\ or infrastructure-level metrics\end{tabular} & \begin{tabular}[c]{@{}l@{}}Improved prediction \\ accuracy\end{tabular} & Static models                                                                                     \\ \hline
\cite{dockercontainer}           & Monolithic                                                                   & Isolation forest                                                                & Classification                                                  & Single cloud            & \begin{tabular}[c]{@{}l@{}}Identification of abnormal \\ resource metrics\end{tabular}                                        & \begin{tabular}[c]{@{}l@{}}Improved detection\\ accuracy\end{tabular}   & Unstable monitoring delays                                                                         \\ \hline
\end{tabular}
\end{table}

Anomaly detection is a critical mechanism for identifying abnormal behaviors in system states or application performance. Table \ref{table:ad} describes the reviewed approaches regarding anomaly detection.

Due to the application-centric and decentralized features of containers, they are more likely to experience a wide range of security threats, which may lead to potential propagation delays in container image dependency management or security attacks \cite{dockerhub}. Tunde-Onadele et al. \cite{containerdetection} classify 28 container vulnerability scenarios into six categories and develop a detection model, including both dynamic and static anomaly detection schemes with KNN, K-means, self-organization map algorithms, which could reach detection coverage up to 86\%.

To balance the energy efficiency and resource utilization of a containerized computing system under hybrid cloud environments, Chhikara et al. \cite{iotcontainer} employ K-means, Hierarchical clustering algorithms, and ensemble learning for identification and classification of underloaded and overloaded hosts. Further container migration operations will be conducted between these two groups for load balancing and energy consumption reduction.

Shah et al. \cite{denpendencyana} extend the LSTM model to analyze the long-term dependencies of real-time performance metrics among microservice units. Relying on the constructed model, they manage to identify critical performance indicators that support anomaly detection of various application and infrastructure-level metrics, including network throughput and CPU utilization.    

\subsubsection{Dependency Analysis}
\label{subsubsection:da}

\begin{table}[h]
\caption{Summary of Dependency Analysis Approaches}
\label{table:wds}
\centering
\tiny
\begin{tabular}{|l|l|l|l|l|l|l|l|}
\hline
\textbf{Ref.}                            & \textbf{\begin{tabular}[c]{@{}l@{}}Application \\ Architecture\end{tabular}} & \textbf{Methods} & \textbf{Type}                                                   & \textbf{Infrastructure} & \textbf{Objectives}                                                                                        & \textbf{Advantages}                                                        & \textbf{Limitations}                                                                                         \\ \hline
\cite{loadbalancemicro} & Microservice                                                                 & BO, GP           & Regression                                                      & Multi-cloud             & \begin{tabular}[c]{@{}l@{}}Discovery of optimal route \\ of individual microservice\end{tabular}           & Load balance                                                               & \begin{tabular}[c]{@{}l@{}}Poor performance under highly\\ dynamic environment\end{tabular}                  \\ \hline
\cite{firm}             & Microservice                                                                 & SVM              & Classification                                                  & Single cloud            & \begin{tabular}[c]{@{}l@{}}Identification of potential \\ heavy-loaded microservice \\ units\end{tabular}  & High accuracy                                                              & \begin{tabular}[c]{@{}l@{}}Implicit explanation of the time \\ overhead and computational costs\end{tabular} \\ \hline
\cite{sinan}            & Microservice                                                                 & CNN              & Classification                                                  & Single cloud            & \begin{tabular}[c]{@{}l@{}}Dependency analysis of \\ microservices between \\ pipelined-tiers\end{tabular} & \begin{tabular}[c]{@{}l@{}}Navigation of \\ system complexity\end{tabular} & Overfitting and misprediction                                                                                \\ \hline
\cite{denpendencyana}   & Microservice                                                                 & LSTM             & \begin{tabular}[c]{@{}l@{}}Time series \\ analysis\end{tabular} & Single cloud            & \begin{tabular}[c]{@{}l@{}}Dependency analysis \\ among microservice units\end{tabular}                    & \begin{tabular}[c]{@{}l@{}}Improved prediction \\ accuracy\end{tabular}    & Ignorance of dependency updates                                                                              \\ \hline
\end{tabular}
\end{table}

Table \ref{table:wds} summarizes the recent studies in service dependency analysis of containerized applications. As serverless function chains or workflows are usually pre-defined by users \cite{serverlessworkflow}, current ML-based dependency analysis solutions only focus on decomposing the internal structures of microservice units and monitoring any dynamic structural updates.

SVM is utilised by Qiu et al. \cite{firm} to find the microservice units with higher risks causing SLO violations through analysis of the performance metrics related to the critical path (CP) of each individual microservice. CP is defined as the longest path between the client request and the underlying microservice in the execution history graph. Since CP can change dynamically at runtime in response to potential resource contention or performance interference, the SVM classifier is implemented with incremental learning for dependency analysis in a dynamic and consistent manner.

Load balancing between microservices under the multi-cloud environment could be rather complex, because of the unstable network latency, dynamic service configuration, and fluctuating application workloads. To address this challenge, Cui et al. \cite{loadbalancemicro} leverage the BO search algorithm with GP to produce the optimal load-balanced request chain for each microservice unit. Then all the individual request chains are consolidated into a tree structure dependency model that can be dynamically updated in case of potential environmental changes.

\subsection{Resource Provisioning}

In this section, we discuss various resource provisioning techniques, including scheduling, scaling, and migration. 

\subsubsection{Scheduling}

\begin{table}[h]
\caption{Summary of Scheduling Approaches}
\label{table:schedule}
\centering
\tiny
\begin{tabular}{|l|l|l|l|l|l|l|l|}
\hline
\textbf{Ref.}                               & \textbf{\begin{tabular}[c]{@{}l@{}}Application \\ Architecture\end{tabular}} & \textbf{Methods}                                             & \textbf{\begin{tabular}[c]{@{}l@{}}Task \\ Structure\end{tabular}} & \textbf{Infrastructure} & \textbf{Objectives}                                                                                   & \textbf{Advantages}                                                                             & \textbf{Limitations}                                                                                           \\ \hline
\cite{fifer}               & Serverless                                                                   & Heuristic                                                    & \begin{tabular}[c]{@{}l@{}}Multiple \\ independent\end{tabular}    & Single cloud            & \begin{tabular}[c]{@{}l@{}}Resource utilization \\ improvement and \\ energy saving\end{tabular}      & \begin{tabular}[c]{@{}l@{}}Reduction of cold start and\\ response latency\end{tabular}         & \begin{tabular}[c]{@{}l@{}}Poor efficiency for tasks\\ with long lifetimes\end{tabular}                        \\ \hline
\cite{edgecloudserverless} & Serverless                                                                   & Heuristic                                                    & Single                                                             & Hybrid cloud            & \begin{tabular}[c]{@{}l@{}}Cost and latency \\ minimization\end{tabular}                              & \begin{tabular}[c]{@{}l@{}}Multi-objective task \\ placement\end{tabular}                       & \begin{tabular}[c]{@{}l@{}}Limited accuracy under\\ high load variance\end{tabular}                            \\ \hline
\cite{cose}                & Serverless                                                                   & Heuristic                                                    & Graph-based                                                        & Hybrid cloud            & Cost minimization                                                                                     & SLA assurance                                                                                   & \begin{tabular}[c]{@{}l@{}}Simplicity of application \\ workloads\end{tabular}                                 \\ \hline
\cite{htas}                & Monolithic                                                                   & Heuristic                                                    & Single                                                             & Single cloud            & \begin{tabular}[c]{@{}l@{}}Resource utilization \\ optimization\end{tabular}                          & Load balance                                                                                    & High scheduling delays                                                                                          \\ \hline
\cite{fitnessaware}        & Monolithic                                                                   & Heuristic                                                    & Single                                                             & Hybrid cloud            & \begin{tabular}[c]{@{}l@{}}Automated task \\ deployment\end{tabular}                                  & \begin{tabular}[c]{@{}l@{}}Optimized container build \\ time and provisioning time\end{tabular} & \begin{tabular}[c]{@{}l@{}}High computational \\ expenses and time \\ overhead\end{tabular}                    \\ \hline
\cite{kuberknots}          & Monolithic                                                                   & Heuristic                                                    & Single                                                             & Single cloud            & \begin{tabular}[c]{@{}l@{}}Resource utilization\\  and energy efficiency \\ optimization\end{tabular} & QoS improvement                                                                                 & \begin{tabular}[c]{@{}l@{}}Insufficient analysis of \\ computational costs and \\ time complexity\end{tabular} \\ \hline
\cite{miras}               & Microservice                                                                 & Actor-Critic                                                 & Graph-based                                                        & Single cloud            & Cost saving                                                                                           & \begin{tabular}[c]{@{}l@{}}Training time reduction \\ and accuracy improvement\end{tabular}     & \begin{tabular}[c]{@{}l@{}}Limitation caused by \\ scarce data\end{tabular}                                    \\ \hline
\cite{video}               & Microservice                                                                 & Heuristic                                                    & Graph-based                                                        & Single cloud            & \begin{tabular}[c]{@{}l@{}}Resource utilization \\ optimization\end{tabular}                          & Cost saving                                                                                     & High scheduling delays                                                                                          \\ \hline
\cite{news}                & Microservice                                                                 & \begin{tabular}[c]{@{}l@{}}Q-Learning, \\ SARSA\end{tabular} & Graph-based                                                        & Single cloud            & \begin{tabular}[c]{@{}l@{}}Minimization of \\ task execution time\end{tabular}                        & SLA assurance                                                                                   & Limited scalability                                                                                            \\ \hline
\cite{jobplacementdmlc}    & Microservice                                                                 & \begin{tabular}[c]{@{}l@{}}Actor-Critic,\\ ANN\end{tabular}  & Single                                                             & Single cloud            & \begin{tabular}[c]{@{}l@{}}Minimization of task\\  completion time\end{tabular}                       & \begin{tabular}[c]{@{}l@{}}Performance interference \\ awareness\end{tabular}                   & \begin{tabular}[c]{@{}l@{}}Implicit description of \\ the space and time \\ complexity\end{tabular}            \\ \hline
\end{tabular}
\end{table}

As shown in Table \ref{table:schedule}, various algorithms have been proposed to solve the scheduling issue for improving system and application performance.

Most of the reviewed approaches follow a design pattern of combining ML-based Workload Modellers or Performance Analyzers with a heuristic scheduling Decision Maker, as discussed in Section \ref{subsubsection:mloe}. As the system complexity has been navigated by prediction models, bin packing and approximation algorithms such as best fit or least fit are commonly adopted to make scheduling decisions with improved resource utilization and energy efficiency \cite{htas,video,edgecloudserverless,fifer, fitnessaware,cose}. For example, Venkateswaran and Sarkar \cite{fitnessaware} manage to significantly reduce the complexity of selecting the best-fit container system and cluster compositions under multi-cloud environments, standing on their prediction model described in Section \ref{subsubsection:bmp}. To mitigate the resource contention between co-located tasks deployed at the same host, Thinakaran et al. \cite{kuberknots} implement a correlation-based scheduler for handling task co-location by measuring GPU consumption correlation metrics, especially consecutive peak resource demand patterns recognized by its ARIMA prediction model. 

The rest of the studies choose RL models as the core component in their scheduling engine. Orhean et al. \cite{news} choose two classic model-free RL models, namely Q-Learning and SARSA, to schedule a group of tasks in a DAG structure. To reduce the overall DAG execution time, the internal tasks in a DAG categorized by their key features and priorities are scheduled under consideration of the dynamic cluster state and machine performance. To address the limitation of high sample complexity of model-free RL approaches, Zhang et al. \cite{miras} attempt to handle scientific workflows under microservice architectures through model-based RL. An ANN model is trained to emulate the system behavior by identification of key performance metrics collected from the microservice infrastructure, so that the synthetic interactions generated by ANN could directly be involved in the policy training process with Actor-Critical to generate scheduling decisions. In such a way, it simplifies the system model and avoids the time consuming and computationally expensive interactions within the real microservice environment.

In conclusion, heuristic or RL models are applied in most of the investigated works for decision making in task scheduling, to achieve resource utilization improvement and task completion time minimization.

\subsubsection{Scaling}

\begin{table}[h]
\caption{Summary of Scaling Approaches}
\label{table:scale}
\centering
\tiny
\begin{tabular}{|l|l|l|l|l|l|l|l|}
\hline
\textbf{Ref.}                              & \textbf{\begin{tabular}[c]{@{}l@{}}Application \\ Architecture\end{tabular}} & \textbf{Methods}                                                     & \textbf{Mechanism} & \textbf{Infrastructure} & \textbf{Objectives}                                                                                                              & \textbf{Advantages}                                                                                      & \textbf{Limitations}                                                                                           \\ \hline
\cite{functionserverless} & Serverless                                                                   & Heuristic                                                            & Horizontal         & Single cloud            & \begin{tabular}[c]{@{}l@{}}Reduction of execution \\ latency and resource \\ consumption\end{tabular}                            & Alleviation of cold starts                                                                               & \begin{tabular}[c]{@{}l@{}}Instability under changing \\ workloads\end{tabular}                                \\ \hline
\cite{rlserverless}       & Serverless                                                                   & Q-Learning                                                           & Horizontal         & Single cloud            & SLA assurance                                                                                                                    & \begin{tabular}[c]{@{}l@{}}Reduction of time overhead \\of cold starts\end{tabular}                     & Simplicity of training model                                                                                   \\ \hline
\cite{autopilot}          & Monolithic                                                                   & Ensemble                                                             & Hybrid             & Single cloud            & Resource saving                                                                                                                  & \begin{tabular}[c]{@{}l@{}}High resource efficiency \\ and reliability\end{tabular}                      & Limited flexibility                                                                                            \\ \hline
\cite{htas}               & Monolithic                                                                   & Heuristic                                                            & Horizontal         & Single cloud            & \begin{tabular}[c]{@{}l@{}}Resource utilization and SLA\\  assurance\end{tabular}                                                & High resource efficiency                                                                                 & Frequent cluster resizing                                                                                      \\ \hline
\cite{sarsa}              & Monolithic                                                                   & \begin{tabular}[c]{@{}l@{}}SARSA, \\ Q-Learning\end{tabular}         & Horizontal         & Single cloud            & \begin{tabular}[c]{@{}l@{}}Resource utilization \\ optimization\end{tabular}                                                      & SLA violation reduction                                                                                  & Inaccuracy due to cold starts                                                                                  \\ \hline
\cite{hybridscaling}      & Monolithic                                                                   & \begin{tabular}[c]{@{}l@{}}Model-based \\ RL\end{tabular}            & Hybrid             & Single cloud            & \begin{tabular}[c]{@{}l@{}}Minimization of application \\ performance penalty, \\ adaption costs, and resource \\ usage\end{tabular} & Improved training speed                                                                                  & \begin{tabular}[c]{@{}l@{}}Simplicity of application \\ models\end{tabular}                                    \\ \hline
\cite{Geocontainer}       & Monolithic                                                                   & \begin{tabular}[c]{@{}l@{}}Model-based \\ RL, heuristic\end{tabular} & Hybrid             & Multi-cloud             & \begin{tabular}[c]{@{}l@{}}Optimal application \\ performance and adaption \\ time\end{tabular}                                 & Cost saving                                                                                              & \begin{tabular}[c]{@{}l@{}}Simplicity of application \\ structures and QoS \\ requirements\end{tabular}        \\ \hline
\cite{FScaler}            & Monolithic                                                                   & MDP, SARSA                                                           & Horizontal         & Hybrid Cloud            & \begin{tabular}[c]{@{}l@{}}Minimization of resource \\ consumption and response \\ time\end{tabular}                              & High scalability                                                                                         & Limited dimensionality                                                                                         \\ \hline
\cite{microsacler}        & Microservice                                                                 & BO, GP                                                               & Horizontal         & Single cloud            & SLA assurance                                                                                                                    & \begin{tabular}[c]{@{}l@{}}High precision and short \\ training time\end{tabular}                        & \begin{tabular}[c]{@{}l@{}}Poor performance and \\ sub-optimal decisions under \\ workload spikes\end{tabular} \\ \hline
\cite{mlmicro}            & Microservice                                                                 & RF                                                                   & Horizontal         & Single cloud            & SLA assurance                                                                                                                    & \begin{tabular}[c]{@{}l@{}}Container expansion time \\ reduction\end{tabular}                            & \begin{tabular}[c]{@{}l@{}}Limited QoS and workload \\ scenarios\end{tabular}                                  \\ \hline
\cite{negotiation}        & Microservice                                                                 & Q-Learning                                                           & Vertical           & Single cloud            & \begin{tabular}[c]{@{}l@{}}Optimization of resource \\ configurations and costs\end{tabular}                                & SLA assurance                                                                                            & Limited workload scenarios                                                                                     \\ \hline
\cite{highutiautoscaling} & Microservice                                                                 & Heuristic                                                            & Horizontal         & Single cloud            & Cost minimization                                                                                                                & \begin{tabular}[c]{@{}l@{}}resource utilization \\ improvement\end{tabular}                              & High time complexity                                                                                           \\ \hline
\cite{mlbasedautoscaling} & Microservice                                                                 & Heuristic                                                            & Horizontal         & Single cloud            & Resource efficiency                                                                                                              & \begin{tabular}[c]{@{}l@{}}Avoidance of oscillations \\ under unexpected workload \\ spikes\end{tabular} & Rigid scaling mechanisms                                                                                       \\ \hline
\cite{Probabilistic}      & Microservice                                                                 & Heuristic                                                            & Horizontal         & Single cloud            & SLA assurance                                                                                                                    & \begin{tabular}[c]{@{}l@{}}Resource utilization \\ optimization\end{tabular}                             & High time complexity                                                                                           \\ \hline
\cite{aiautoscale}        & Microservice                                                                 & Heuristic                                                            & Horizontal         & Single cloud            & SLA assurance                                                                                                                    & Lower request loss                                                                                       & High resource usage                                                                                            \\ \hline
\cite{slocloudnative}     & Microservice                                                                 & Heuristic                                                            & Vertical           & Single cloud            & SLO assurance                                                                                                                    & \begin{tabular}[c]{@{}l@{}}Resource utilization \\ optimization\end{tabular}                             & \begin{tabular}[c]{@{}l@{}}Simplicity of application \\ datasets\end{tabular}                                  \\ \hline
\cite{HANSEL}             & Microservice                                                                 & SARSA                                                                & Horizontal         & Hybrid cloud            & \begin{tabular}[c]{@{}l@{}}Resource utilization \\ optimization\end{tabular}                                                     & SLA assurance                                                                                            & \begin{tabular}[c]{@{}l@{}}Capability limitation of edge\\devices\end{tabular}                               \\ \hline
\cite{sinan}              & Microservice                                                                 & Heuristic                                                            & Horizontal         & Single cloud            & QoS assurance                                                                                                                    & \begin{tabular}[c]{@{}l@{}}Resource utilization \\ optimization\end{tabular}                             & \begin{tabular}[c]{@{}l@{}}Implicit evaluation of \\ computational costs and \\ time complexity\end{tabular}   \\ \hline
\cite{firm}               & Microservice                                                                 & Actor-critic                                                         & Hybrid             & Single cloud            & \begin{tabular}[c]{@{}l@{}}Resource utilization \\ optimization\end{tabular}                                                     & SLO violation mitigation                                                                                 & \begin{tabular}[c]{@{}l@{}}Limited scalability and \\ anomaly detection\end{tabular}                           \\ \hline
\end{tabular}
\end{table}

Scaling could dynamically adjust the system states in response to the changing workloads and cloud environments. The research works related to scaling are given in Table \ref{table:scale}.

To improve the decision quality and accuracy of RL-based autoscalers for monolithic applications with SLA assurance, Zhang et al. \cite{sarsa} build a horizontal scaling approach through the SARSA algorithm, based on analysis of the application workloads and CPU usage predicted by the ARIMA and ANN models. For addressing the same research questions under fog-cloud environments, Sami et al. \cite{FScaler} simulate the horizontal scaling of containers as an MDP model that considers both the changing workloads and free resource capacity in fogs. Then SARSA is chosen for finding the optimal scaling strategy on top of the MDP model through online training at a small data scale. 

Further considering the possibility of hybrid scaling of monolithic applications, Rossi et al. \cite{hybridscaling} propose a model-based RL approach, targeting to find a combination of horizontal and vertical scaling decisions that meet the QoS requirements with the lowest adaption costs and resource wastage. Furthermore, the authors extend this model with a network-aware heuristic method for container placement in geographically distributed clouds \cite{Geocontainer}. Although the approaches mentioned above can improve resource utilization and QoS to a certain degree, their application workloads and QoS scenarios are too simple, without enough consideration of the diversity and complexity of cloud workloads.

Plenty of previous studies \cite{highutiautoscaling,mlbasedautoscaling,aiautoscale,Probabilistic,slocloudnative} have tried to leverage heuristic methods for microservice scaling, assisted by ML-based workload modelling and performance analysis. However, such approaches underestimate the inter-dependencies between microservices that are updated dynamically. On the other hand, model-based RL algorithms are usually unsuitable for microservice-based applications for the same reason. As microservice dependencies could potentially change at runtime, the simulation of state transitions could then be invalid. 

Therefore, model-free RL algorithms are more common in the application scaling of microservices, as they do not rely on transition models. Qiu et al. \cite{firm} implement an SLO violation alleviation mechanism using Actor-Critic to scale the critical microservices detected by SVM as mentioned in Section \ref{subsubsection:da}. The Actor-Critic model produces a horizontal scaling decision to minimize the SLO violations through evaluating three crucial features, including SLO maintenance ratio, workload changes, and request composition. To speed up the model training process of RL methods, Yan et al. \cite{HANSEL} design a multi-agent parallel training model based on SARSA for horizontal scaling of microservices under hybrid clouds. Assisted with the workload prediction results generated by Bi-LSTM , their elastic scaling approach could make a more accurate scaling decision of when, where, and how many microservice instances should be scaled up/down. In such a way, it achieves significant resource utilization improvement and cost reduction under SLA assurance. 

Cold starts during the invocation of serverless functions are a serious performance bottleneck of serverless computing platforms. Cold starts are defined as the time overhead of environment setup and function initialization. Xu et al. \cite{functionserverless} present a container pool scaling strategy, where function containers are pre-initialized by evaluating the first function invocation time (predicted by LSTM as discussed in Section \ref{subsubsection:bmp}) and the number of containers in each function category. Similarly, Agarwal et al. \cite{rlserverless} introduce a Q-Learning agent to summarize the function invocation patterns and make the optimal scaling decisions of function containers in advance. A series of metrics, including the number of available function containers, per-container CPU utilization, and success/failure rates, are selected to represent the system states in the Q-Learning model. Due to the nature of model-free RL methods, prior information of the input function is not necessary.

In summary, there has been a large body of literature in the area of scaling using diverse solutions, covering all types of application architectures and cloud infrastructures. The majority of the reviewed works are related to autoscaling of microservice-based applications, with the model-free RL models as the latest resolution to process the dynamically changing workloads and microservice dependencies \cite{negotiation,HANSEL,firm}. Their common targets mainly focus on SLA assurance and resource efficiency optimization.

\subsubsection{Migration}

\begin{table}[h]
\caption{Summary of Migration Approaches}
\label{table:migra}
\centering
\tiny
\begin{tabular}{|l|l|l|l|l|l|l|l|}
\hline
\textbf{Ref.}                                    & \textbf{\begin{tabular}[c]{@{}l@{}}Application \\ Architecture\end{tabular}} & \textbf{Methods}                                                & \textbf{Mechanism} & \textbf{Infrastructure} & \textbf{Objectives}                                                                            & \textbf{Advantages}                                                        & \textbf{Limitations}                                                               \\ \hline
\cite{htas}                     & Monolithic                                                                   & Heuristic                                                       & Rescheduling       & Single cloud            & SLA assurance                                                                                  & \begin{tabular}[c]{@{}l@{}}Resource contention \\ alleviation\end{tabular} & Long execution delays                                                               \\ \hline
\cite{migrationfog}             & Monolithic                                                                   & \begin{tabular}[c]{@{}l@{}}MDP, DNN, \\ Q-Learning\end{tabular} & Offloading         & Hybrid cloud            & \begin{tabular}[c]{@{}l@{}}Reduction of communication \\ delays, power consumption\end{tabular} & Low migration costs                                                        & High space complexity                                                              \\ \hline
\cite{iotcontainer}             & Monolithic                                                                   & Heuristic                                                       & Offloading         & Hybrid cloud            & Energy efficiency improvement                                                                  & Load balance                                                               & \begin{tabular}[c]{@{}l@{}}High time and space \\ complexity\end{tabular}          \\ \hline
\cite{video}                    & Microservice                                                                 & Heuristic                                                       & Rescheduling       & Single cloud            & Resource utilization optimization                                                              & Cost saving                                                                & SLA violations                                                                     \\ \hline
\cite{microservicecoordination} & Microservice                                                                 & \begin{tabular}[c]{@{}l@{}}Q-Learning,\\ MDP\end{tabular}       & Offloading         & Hybrid cloud            & \begin{tabular}[c]{@{}l@{}}Reduction of service delays and \\ migration cost\end{tabular}       & Optimized performance                                                      & \begin{tabular}[c]{@{}l@{}}Lack of consideration on \\ load balancing\end{tabular} \\ \hline
\end{tabular}
\end{table}

As depicted in Table \ref{table:migra}, Migration is a complementary mechanism for load balancing across cloud environments.

Some reviewed approaches manage to bind ML-based behavior models with heuristic migration algorithms. Zhong et al. \cite{htas} develop a least-fit rescheduling algorithm to evict and relocate a set of lower-priority containers with the least QoS impact when unexpected resource contention occurs between co-located containers. The rescheduling algorithm readjusts the container configuration based on runtime performance metrics and selects the node with the most available resources evaluated by K-means for relocation. Besides, Chhikara et al. \cite{iotcontainer} introduce an energy-efficient offloading model with a set of heuristic methods, including random placement, first-fit, best-fit, and correlation threshold-based placement algorithms. It is aimed at resource load balancing under hybrid cloud environments by migrating containers from overloaded nodes to underloaded nodes that are identified through classification as described in Section \ref{subsubsection:ad}. However, these approaches may come with high execution delays in large-scale computing systems.

A fog-cloud container offloading prototype system is presented by Tang et al. \cite{migrationfog}. The container offloading process is considered as a multi-dimensional MDP model. To reduce the network delays and computation costs under potentially unstable environments, the deep Q-Learning algorithm combines the deep neural network (DNN) and Q-Learning model to quickly produce an efficient offloading plan. Wang et al. \cite{microservicecoordination} further extends the combination of MDP and Q-Learning in the context of microservice coordination under edge-cloud environments. The process of microservice coordination is assumed as a sequential decision scheme and formulated as an MDP model. On top of the MDP model, Q-Learning is used to find the optimal solution for service migration or offloading,  in light of long-term performance metrics, including overall migration delays and costs. 

\subsection{Summary}
In this section, we summarize the discussed researches based on three important categories, namely application architecture, infrastructure, and hybrid machine learning model. 

\subsubsection{Application Architecture}

\begin{figure}[h]
 \centering
 \includegraphics[width=0.3\linewidth]{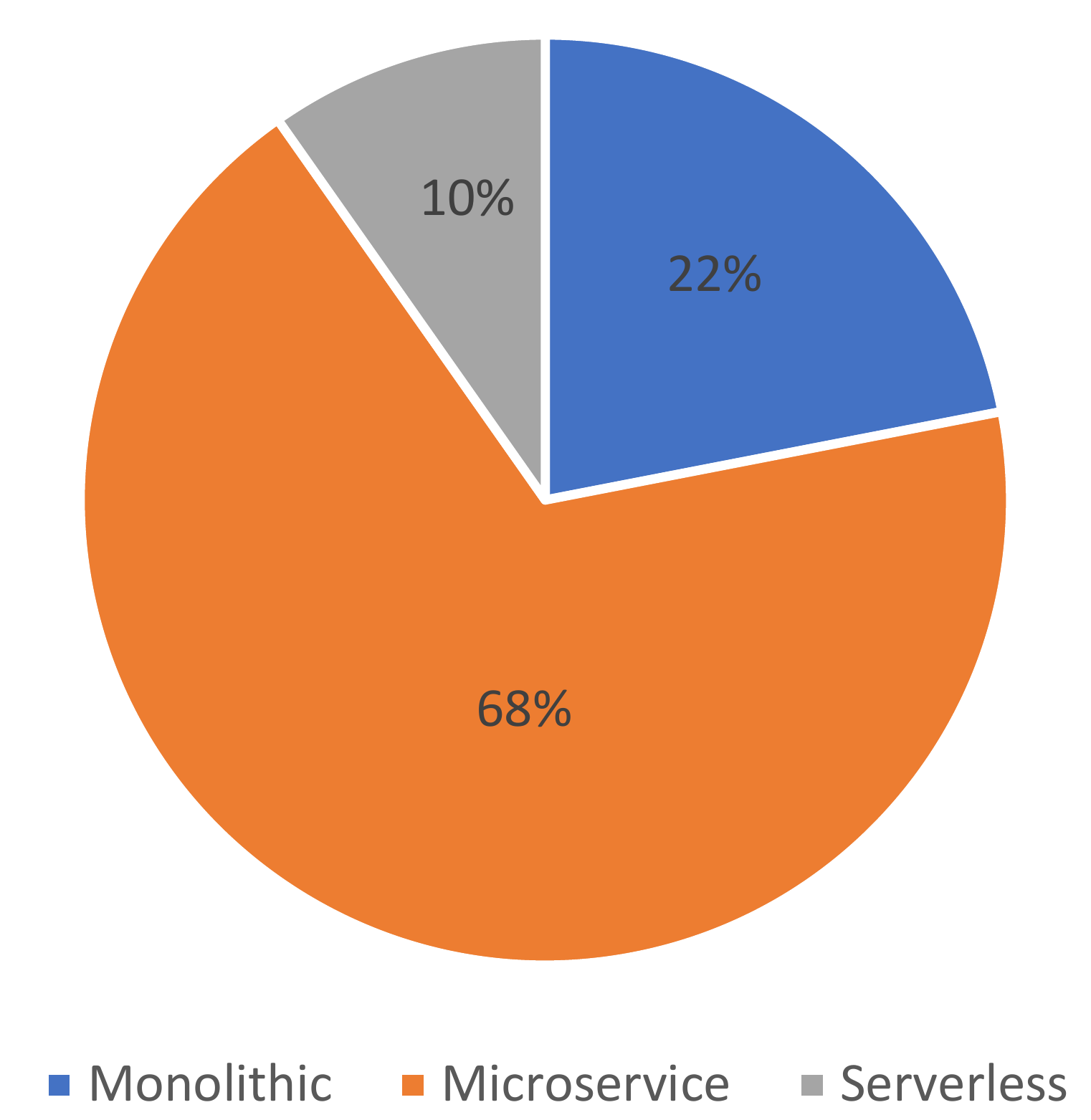}
 \caption{Study Distribution of Application Architectures}
 \Description{Application Archi}
 \label{fig:aa}
\end{figure}

As described in Fig. \ref{fig:aa}, most of the reviewed papers (68\%) are concerned with modelling and management of microservice-based applications. In light of the dynamic and decentralized nature of microservices, diverse ML algorithms have been investigated for capturing the key characteristic of microservice workloads, performance, and inter-dependencies, such as BGR, LSTM, GRU, SVM, GP, ANN, CNN, and DT \cite{highutiautoscaling,mlbasedautoscaling,gruprediction, miras,loadbalancemicro,sinan, denpendencyana}. As the inter-dependencies of microservice units could dynamically update at runtime, the key concern of microservice resource provisioning is the chain reactions caused by microservice unit scaling operations. By combining ML-based behavior models with heuristic or model-free RL methods for scaling, the most recent studies manage to achieve high resource utilization optimization, cost reduction, and SLA assurance \cite{highutiautoscaling,mlbasedautoscaling,aiautoscale,Probabilistic,slocloudnative,firm,HANSEL}.

The research works related to the orchestration of single-component containerized applications hold a study distribution of 22\%. As the workload scenarios of these applications are relatively simple, their core drive of workload/behavior modelling is to predict their resource demands under certain QoS requirements. Assisted with such prediction results, heuristic and RL methods (both model-free and model-based) are adopted for optimization of the resource allocation process regarding cost, energy, and resource efficiency \cite{htas,sarsa,hybridarima,hybridworkloadprediction, fitnessaware,kuberknots,performancedocker}.

Although the serverless architecture is currently having the lowest study distribution (10\%), it is enjoying a growing popularity and becoming a prevalent application architecture in cloud computing. Most of the existing ML-based researches in this field are trying to alleviate the performance downgrade caused by function cold starts. LSTM and GBR models have been employed to estimate the function invocation time in serverless function chains, while the allocation of functions and scaling of function containers are mainly solved by heuristic methods and Q-Learning \cite{edgecloudserverless,functionserverless,fifer,rlserverless}.

\subsubsection{Infrastructure}

\begin{figure}[h]
 \centering
 \includegraphics[width=0.35\linewidth]{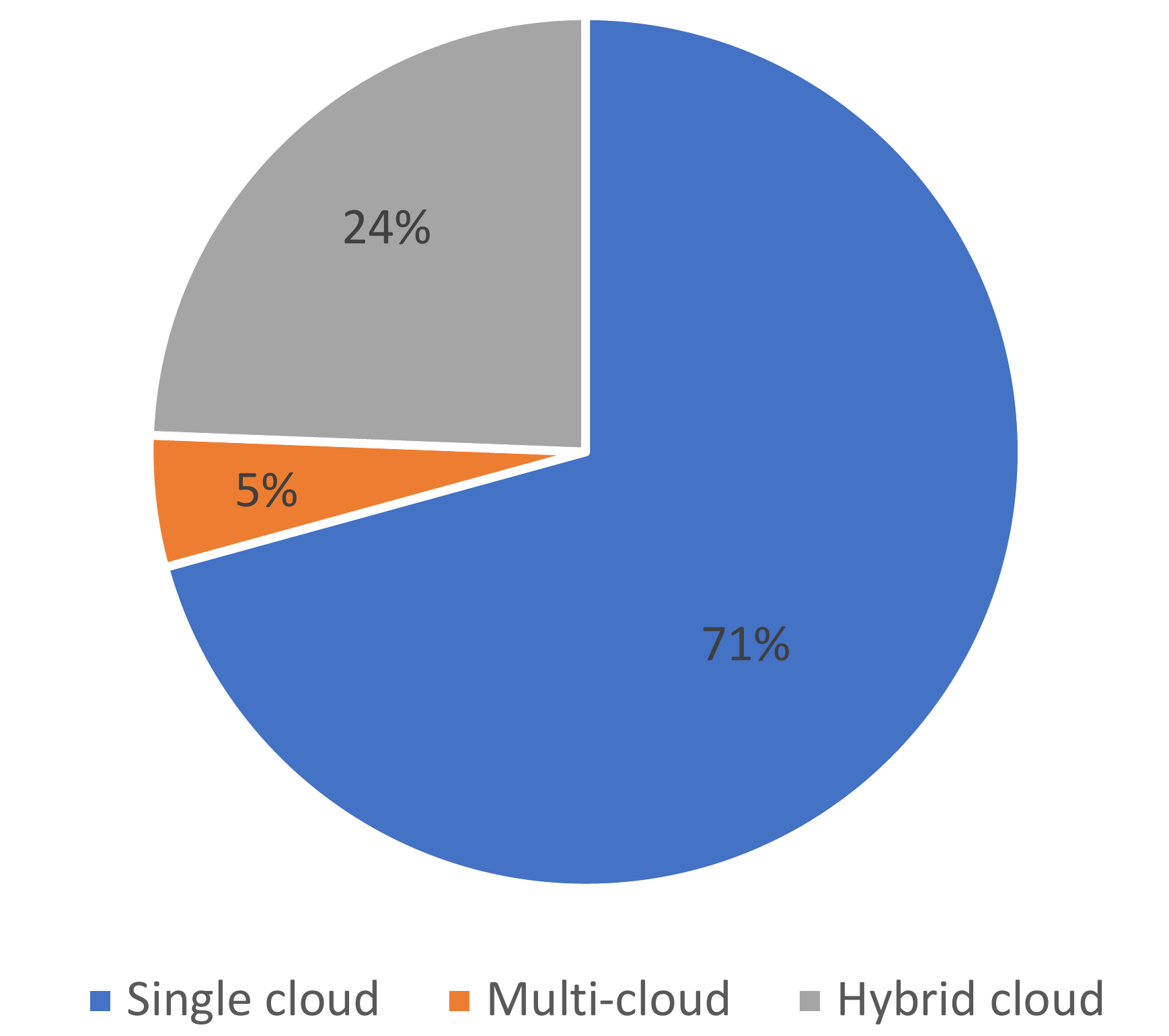}
 \caption{Study Distribution of Cloud Infrastructures}
 \Description{Cloud Infra}
 \label{fig:ci}
\end{figure}

The majority of the included researches (71\%) only consider application deployment under single cloud environments as demonstrated in Fig. \ref{fig:ci}. Since single cloud environments are easier to manage, it is not necessary to raise the system complexity for most applications. Only 5\% of studies have attempted behavior modelling and scaling under geographic-distributed multi-cloud environments \cite{loadbalancemicro,Geocontainer}, because the interactions under such context are usually time-consuming and computation-intensive.

As traditional cloud computing commonly causes significant propagation delays, bandwidth and energy consumption by hosting all the applications and data in cloud servers, edge and fog computing is emerging as mainstream computing paradigms. Therefore, the latest studies have investigated the possibility of container orchestration under hybrid cloud environments, where the crucial research question is to decide where to host the containerized applications among cloud/edge devices and the cloud. The proposed solutions for scheduling and offloading of containerized applications under hybrid clouds, including MDP, RL, DRL and heuristic methods \cite{edgecloudserverless,fitnessaware,FScaler,HANSEL,migrationfog,iotcontainer}, aim to reduce end-to-end latency and optimize resource/energy efficiency.

\subsubsection{Hybrid Machine Learning Model} 

\begin{figure}[h]
 \centering
 \includegraphics[width=0.65\linewidth]{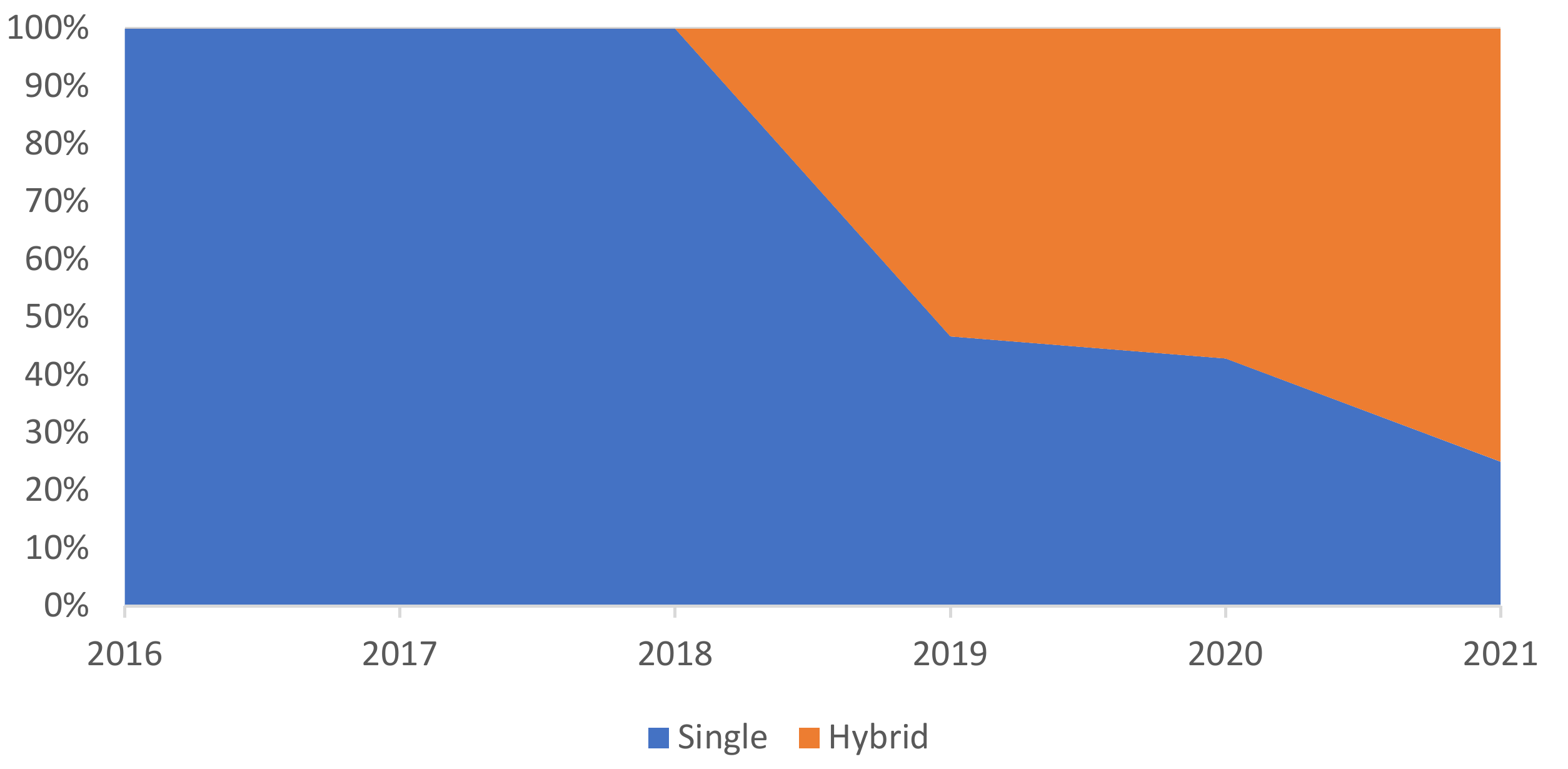}
 \caption{Study Distribution of Machine Learning Models between 2016 and 2021}
 \Description{Machine Learning Model}
 \label{fig:mlm}
\end{figure}

Fig. \ref{fig:mlm} shows the study distribution of ML models between 2016 and 2021, where we can observe a rising trend of hybrid ML models. A single ML model consisting of merely one ML algorithm is designed to solve a specific container orchestration problem, either data analysis or resource provisioning. By 2018, only single ML models had been adopted in this field. As the internal structures of containerized applications like microservices are becoming rather complex and dynamic with continuously growing demands of higher modelling/prediction accuracy and lower response time, this requires ML models to be more robust and efficient with even lower error rates, computation costs, and data training time. Therefore, most studies have been investigating hybrid ML models composed of a mixture of multiple ML algorithms to solve one or more orchestration problems from 2019  \cite{hybridarima,hybridworkloadprediction,gruesprediction,gruprediction,microsacler,migrationfog,microservicecoordination}.

\section{Future Directions}
\label{section:future}
Although the existing studies have covered diverse orchestration schemes, application architectures, and cloud infrastructures, some research gaps and challenges are only partially addressed. In this section, we discuss a series of open research opportunities and potential future directions :

\begin{enumerate}
    \item \textbf{Workload Distribution in Microservices.} The current workload characterization methods are mainly focusing on modelling and prediction of the request arrival rates and resource usage patterns. Very few works try to address the issue of workload distribution across microservice units. The changing workload distribution on individual microservices could potentially cause a chain reaction and further impact the overall application performance. Therefore, how to simulate and standardize workload distribution between microservices for load balance and performance optimization remains an unsolved research question.
    
    \item \textbf{Microservice Dependency Analysis.} The dynamic inter-dependencies between microservices is a crucial part of application complexity navigation. Though some works have attempted to predict application performance or identity critical microservice units through dependency analysis, there is no existing solution to explicitly address the relationship between the status of individual microservices and overall application performance metrics. Such analysis models are necessary for scheduling and scaling microservice-based applications for performance optimization and SLA assurance. 
    
    \item \textbf{Systematic Anomaly Detection and Recovery.} The current ML-based anomaly detection methods for containerized applications are mostly based on resource/performance metrics or security threats. There is a need for a systematic approach for anomaly detection, root cause categorization, and recovery in a timely and accurate fashion, under different application architectures and cloud infrastructures. 
    
    \item \textbf{Graph-based Task Scheduling.} Batch processing jobs consisting of a group of dependent tasks in DAG structures are common in containerized applications, but the literature related to the scheduling problem of such applications is very limited. Some previous studies manage to resolve this problem under a simplified condition of homogeneous job structures where each task in a DAG is configured similarly with the same execution time. This kind of assumptions are rather unrealistic considering the complex and heterogeneous nature of DAG jobs. A sophisticated scheduling strategy should not only consider the overall DAG structure and different task configurations (e.g., resource demands, execution time, and replica sizes), but also the runtime performance metrics and fault tolerance.

    \item \textbf{Management of Function Chains in Serverless Architectures.} The latest ML-based studies on serverless architectures are all related to the alleviation of function cold starts, with many other research directions left to be discovered. As functions are submitted without any prior application-level knowledge, it is tricky to implement a robust and efficient solution for workload classification, resource demand estimation, and autoscaling. How to optimize the invocation of function chains in an SLO and cost-aware manner is also a crucial research question, especially under hybrid cloud environments.
    
    \item \textbf{Microservices in Hybrid Clouds.} Under the emerging trend of edge and fog computing, the most recent studies have made efforts to move microservices to edge/fog devices for communication delays reduction and cost saving. However, the extremely heterogeneous and geographically distributed computation resources within hybrid clouds significantly raise the complexity of application and resource management. A resource provisioning solution under such environments must consider the long-term impacts of each microservice placement or adjustment decision, regarding multi-dimensional optimization objectives, such as costs, energy, and end-to-end latency. Therefore, there is great potential for ML-based solutions to simulate the process of scheduling, scaling, and offloading for microservices under hybrid clouds. 
    
    \item \textbf{Energy-aware Resource Management.} Through combing ML-based workload characterization and performance analysis models, the overall energy consumption of a system could be precisely predicted based on the resource utilization of each PM. Accordingly, these insights grant us more options for developing energy-aware orchestration frameworks. For instance, brownout technologies could be utilised to activate/deactivate optional microservices for energy efficiency optimization, according to the predicted trends in resource utilization and SLA violations \cite{brownout}. Besides, resource provisioning of containerized applications could also be conducted in a load balancing way to reduce energy consumption.
    
    \item \textbf{Multi-dimensional Performance Optimization.} One of the essential optimization problems in container orchestration is to manage the trade-off between different performance metrics, especially SLA violations and financial costs. Though some previous studies \cite{fifer,edgecloudserverless,kuberknots} address this issue to a certain degree by balancing several optimization objectives during resource provisioning, a standard performance analysis benchmark should be designed to accurately decompose the relation between a set of pre-defined key performance metrics. Such knowledge could be further applied in resource provisioning to produce optimal orchestration decisions with multiple performance requirements taken into account. 
    
    \item \textbf{Fully Integrated ML-based Optimization Engine.} Considering the long data training time and large data volumes requested by ML models, a partially integrated ML engine enjoys high popularity, where ML models are only utilised for behavior modelling to assist the original Container Orchestrator in online resource provisioning. Following the reference architecture of ML-based optimization engine in Section \ref{subsubsection:mloe}, a fully integrated ML engine should be capable of combining multiple ML models for both offline training of behavior models and fast online decision making.
    
    \item \textbf{Edge Intelligence.} Due to the extreme data scales and geographical distribution of edge devices, data transition to a centralized ML-based Optimization Engine across edge networks is potentially time-consuming and expensive. Hence, the next-generation edge intelligence agents should follow a decentralized architecture where each independent agent is deployed at a distributed location close to end users for data collection, analytical model construction, and maintenance of serverless or microservice-based applications under low propagation delays.
    
    \end{enumerate}

\section{Conclusion}
\label{section:conclude}

In this work, we have conducted an extensive literature review on how container orchestration is managed using diverse machine learning-based approaches in the state-of-the-art studies, with emphasis on the specific application architectures and cloud infrastructures. The last few years have witnessed a continuously growing trend of the adoption of machine learning methods in containerized application management systems for behavior modelling, as well as resource provisioning of complicated decentralized applications. Compared with most traditional heuristic methods, ML-based models could produce more accurate orchestration decisions with shorter computation delays, under complex and dynamic cloud environments consisting of highly heterogeneous and geographically distributed computation resources.

We have observed machine learning-based methods applied under a wide range of scenarios, but there is no systematic approach to build a complete machine learning-based optimization framework in charge of the whole orchestration process. Under the emerging trend of moving microservice and serverless applications to hybrid clouds, such frameworks are necessary to handle the sustainably growing system complexity and balance multi-dimensional optimization objectives.

\bibliographystyle{unsrt}
\bibliography{./taxonomy}

\end{document}